\documentclass[acmtog]{acmart}

\acmSubmissionID{484}

\usepackage{booktabs} % For formal tables

\usepackage{hyperref}
\usepackage{array}
\usepackage{subcaption}
\usepackage{multirow}
\usepackage[ruled]{algorithm2e} % For algorithms

\SetAlFnt{\small}
\SetAlCapFnt{\small}
\SetAlCapNameFnt{\small}
\SetAlCapHSkip{0pt}

\acmJournal{TOG}

\renewcommand{\paragraph}[1]{{\vspace{1mm}\noindent \bf #1}.}

\copyrightyear{2024}
\acmYear{2024}
\setcopyright{rightsretained}
\acmConference[SA Conference Papers '24]{SIGGRAPH Asia 2024 Conference Papers}{December 3--6, 2024}{Tokyo, Japan}
\acmBooktitle{SIGGRAPH Asia 2024 Conference Papers (SA Conference Papers '24), December 3--6, 2024, Tokyo, Japan}\acmDOI{10.1145/3680528.3687613} \acmISBN{979-8-4007-1131-2/24/12}
\begin{document}
% Title portion
\title{HyperGAN-CLIP: A Unified Framework for Domain Adaptation, Image Synthesis and Manipulation}

% Authors
\author{Abdul Basit Anees}
\email{abdulbasitanees98@gmail.com}
\orcid{0000-0003-1293-1796}
\affiliation{%
  \institution{Koç University}
  \country{Turkey}
}

\author{Ahmet Canberk Baykal}
\email{canberk.baykal1@gmail.com}
\orcid{0000-0002-0249-5858}
\affiliation{%
  \institution{University of Cambridge}
  \country{United Kingdom}
}

\author{Muhammed Burak Kizil}
\email{mkizil19@ku.edu.tr}
\orcid{0009-0008-9007-9280}
\affiliation{%
  \institution{Koç University}
  \country{Turkey}
}

\author{Duygu Ceylan}
\email{duygu.ceylan@gmail.com}
 \orcid{0000-0002-2307-9052}
\affiliation{%
  \institution{Adobe Research}
  \country{United Kingdom}
}

\author{Erkut Erdem}
\email{erkut@cs.hacettepe.edu.tr}
\orcid{0000-0002-6744-8614}
\affiliation{%
  \institution{Hacettepe University}
  \country{Turkey}
}

\author{Aykut Erdem}
\email{aerdem@ku.edu.tr}
\orcid{0000-0002-6280-8422}
\affiliation{%
  \institution{Koç University}
  \country{Turkey}
}

\renewcommand{\shortauthors}{Anees et al.}

\begin{abstract}
Generative Adversarial Networks (GANs), particularly StyleGAN and its variants, have demonstrated remarkable capabilities in generating highly realistic images. Despite their success, adapting these models to diverse tasks such as domain adaptation, reference-guided synthesis, and text-guided manipulation with limited training data remains challenging. Towards this end, in this study, we present a novel framework that significantly extends the capabilities of a pre-trained StyleGAN by integrating CLIP space via hypernetworks. This integration allows dynamic adaptation of StyleGAN to new domains defined by reference images or textual descriptions. Additionally, we introduce a CLIP-guided discriminator that enhances the alignment between generated images and target domains, ensuring superior image quality. Our approach demonstrates unprecedented flexibility, enabling text-guided image manipulation without the need for text-specific training data and facilitating seamless style transfer. Comprehensive qualitative and quantitative evaluations confirm the robustness and superior performance of our framework compared to existing methods.

\end{abstract}




\begin{teaserfigure}
\centering
  \includegraphics[width=\linewidth]{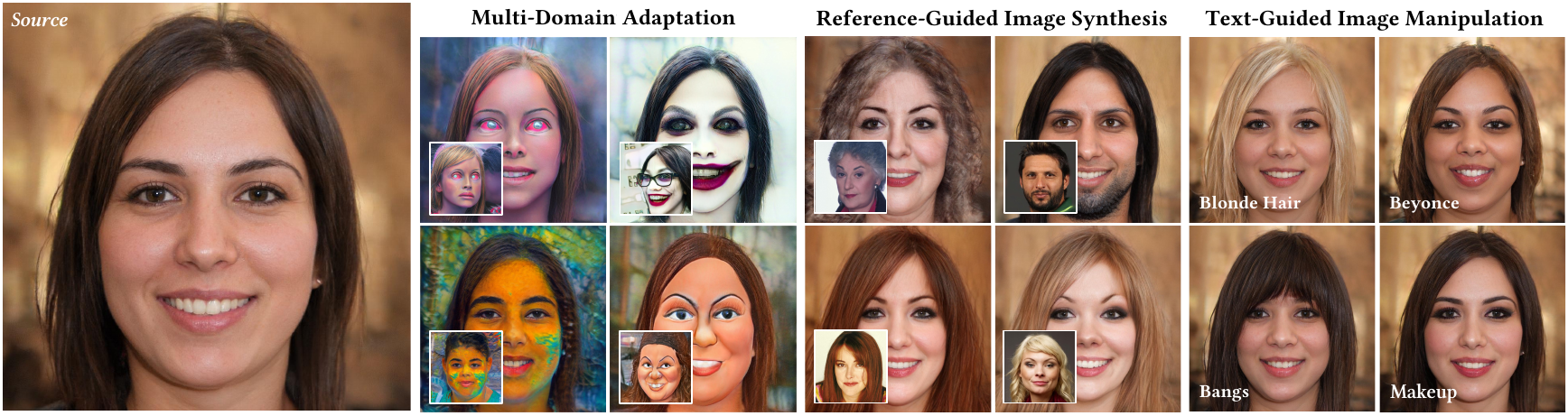}
  \caption{\textbf{HyperGAN-CLIP and its Applications.} Introducing HyperGAN-CLIP, a flexible framework that enhances the capabilities of a pre-trained StyleGAN model for a multitude of tasks, including multiple domain one-shot adaptation, reference-guided image synthesis and text-guided image manipulation. Our method pushes the boundaries of image synthesis and editing, enabling users to create diverse and high-quality images with remarkable ease and precision.}
  \label{fig:teaser}
\end{teaserfigure}

\maketitle

\section{Introduction}
Generative Adversarial Networks (GANs)~\cite{NIPS2014_5ca3e9b1} have dramatically advanced the synthesis of highly realistic images through novel ideas such as progressive growth~\cite{karras2018progressive} and style-based generators~\cite{Karras_2019_CVPR,Karras_2020_CVPR,Karras2021}. These techniques enable the training of cutting-edge GANs on large, high-resolution datasets by exploiting semantically rich latent spaces for precise style manipulation. However, their reliance on substantial training and large datasets poses significant challenges in data-scarce environments. 

Addressing the data scarcity issue, traditional domain adaptation techniques for GANs typically involve fine-tuning pre-trained generators with limited samples from the target domain. While these methods enhance model applicability, they often struggle with a trade-off between the fidelity of domain-specific attributes and the quality of images generated from the source domain. Additionally, methods that utilize multi-modal CLIP embeddings for guided image generation and manipulation \cite{gal2021stylegannada, zhu2021mind} are constrained by the attributes present during training \cite{wei2022hairclip,Lyu_2023_CVPR, CLIPInverter}, and they face difficulties with out-of-distribution images. Per-edit optimization techniques \cite{Patashnik_2021_ICCV, xia2021tedigan, chefer2021targetclip}, though highly flexible, incur substantial computational costs at inference.

In response to these challenges, we propose HyperGAN-CLIP, a~unified framework that not only addresses the limitations of existing domain adaptation methods but also expands their functionality to include reference-guided image synthesis and text-guided image manipulation. This comprehensive framework utilizes a single example from each target domain to efficiently adapt pre-trained GAN models, eliminating the need for task-specific models. Central to HyperGAN-CLIP is a conditional hypernetwork that dynamically adjusts the generator's weights based on domain-specific embeddings from images or text, facilitated by CLIP embeddings.

The strategic use of our hypernetwork module design results in a duplicated generator network that produce domain-specific features via CLIP embeddings. These features are seamlessly integrated into the original generator through a residual feature injection mechanism, which not only preserves the identity of the source domain but also enhances the robustness of the generator by preventing mode collapse. This mechanism effectively addresses common challenges in domain adaptation, and enables our framework to adapt to different domains without requiring separate training sessions for each one. Unlike prior approaches, CLIP-oriented hypernetworks effectively understand and leverage the common characteristics shared among target domains during adaptation, leading to improved results. Moreover, they enhance our framework’s capabilities by allowing the use of images and text prompts as guidance, making it well-suited for tasks like reference-guided image synthesis and text-guided image manipulation. 

In summary, the key contributions of our work are as follows:
\begin{itemize}
    \item We propose a conditional hypernetwork that effectively adapts a pre-trained StyleGAN generator to multiple domains with minimal data, maintaining high-quality synthesis image synthesis without increasing model size.
    \item Our novel design offers more flexibility and supports a wide range of synthesis and editing tasks, including reference-guided image synthesis and text-guided manipulation, without any need for training separate models for each task.
    \item We conduct extensive evaluations across multiple domains and datasets, demonstrating our framework's effectiveness and adaptability compared to existing methods.
\end{itemize}

Our code and models are publicly available at the project website: {\url{https://cyberiada.github.io/HyperGAN-CLIP}}.

%\revised{Our code and models are available at the project website: {\url{URL_IS_HIDDEN}}.}
\section{Related Work}
\subsection{State-of-the-art in GANs}
Field of image synthesis and editing has experienced significant advances through the use of generative adversarial networks (GANs) \cite{NIPS2014_5ca3e9b1}. These advances have been by innovative architectural and training strategies that yield highly realistic images. Notably, PGGAN~\cite{karras2018progressive} introduces progressive resolution enhancement, while BigGAN~\cite{brock2018large} scales up image synthesis with larger batch sizes and introduces techniques like residual connections and the truncation trick for improved quality. StyleGAN~\cite{Karras_2019_CVPR} and its successors, StyleGAN2~\cite{Karras_2020_CVPR} and StyleGAN3~\cite{Karras2021}, further enhance photorealism and reduce artifacts by using a generator inspired by style transfer literature~\cite{Gatys2015ANA}. StyleSwin~\cite{zhang2021styleswin} and GANformer~\cite{hudson2021ganformer} incorporate transformers or bipartite structures to generate complex images with multiple objects.

StyleGAN is particularly acclaimed for its rich, semantically meaningful latent space, which enables users to finely manipulate image attributes. GAN inversion, a common technique to embed real images into this space, can be accomplished through methods such as direct optimization~\cite{CreswellB16b,Abdal2019,abdal2020,Tewari2020}, learning-based approaches~\cite{zhu2020indomain,alaluf2021restyle,Bau2019,richardson2021encoding,Tov2021,bai2022high}, or hybrids~\cite{zhu2016generative,Bau2019Large}. These techniques allow for exploration and manipulation of the latent space to discover and apply meaningful editing directions, often in an unsupervised manner~\cite{voynov2020unsupervised,ganspace2020,shen2021closedform}, or by leveraging image-level attributes~\cite{shen2020interpreting, Abdal2021, wu2020stylespace}.

\subsection{Domain Adaptation for GANs}
Few-shot GAN domain adaptation involves adjusting pre-trained models to new image domains with limited data, often leading to challenges such as overfitting and mode collapse. To address these challenges, several novel strategies have been implemented. \citet{ojha2021few-shot-gan} employ a cross-domain distance consistency loss to maintain diversity while transferring to new domains. \citet{back2021} fine-tunes StyleGAN2 by freezing initial style blocks and adding a structural loss to minimize deviations between the source and target domains. DualStyleGAN~\cite{yang2022Pastiche} employs distinct style paths for content and portrait style transfer, while RSSA~\cite{Xiao_2022_CVPR} compresses the latent space for better domain alignment. StyleGAN-NADA~\cite{gal2021stylegannada} uses CLIP embeddings for directional guidance during adaptation, enhancing the fidelity of transfers. Mind-the-Gap~\cite{zhu2021mind} introduces regularizers to reduce overfitting. JoJoGAN~\cite{chong2022jojo} learns a style mapper from a single example using GAN inversion and StyleGAN's style-mixing property. DiFa~\cite{zhang2022difa} leverages CLIP embeddings for both global and local-level adaptation, and employs selective cross-domain consistency to maintain diversity. OneshotCLIP~\cite{kwon2023tpami} employs a two-step training strategy involving CLIP-guided latent optimization and generator fine-tuning with a novel loss function to ensure CLIP space consistency. DynaGAN~\cite{Kim2022DynaGAN} modulates the pre-trained generator's weights for dynamic adaptation. HyperDomainNet~\cite{alanov2023hyperdomainnet} employs hypernetworks to predict weight modulation parameters, combined with regularizers and a CLIP directional loss for multi-domain adaptation. Adaptation-SCR~\cite{Liu2023scr} proposes a spectral consistency regularizer to alleviate mode collapse and preserve diversity and granularity adaptive regularizer to balance diversity and stylization during domain adaptation. Our method extends these studies by using a hypernetwork to modulate a StyleGAN2 generator's weights, integrating missing domain-specific features into a frozen generator for better identity preservation and minimal distortion. Unlike the direct tuning in DynaGAN, our approach uses CLIP embeddings to generate and inject features, significantly differing from StyleGAN-NADA's finetuning approach, which risks overfitting. Moreover, our hypernetwork is conditioned on multimodal CLIP embeddings, broadening our model's application from domain adaptation to reference-guided image synthesis and text-guided manipulation. %We also implement a novel CLIP-conditioned discriminator to further enhance output quality.

\subsection{Reference-Guided Image Synthesis}
Reference-guided image synthesis combines the content of one image with the style of another, a process that has evolved significantly from early neural style transfer techniques like~\cite{Gatys2015ANA}, which often suffered from style-artifacts due to inadequate handling of local semantic details. To improve upon these limitations, WCT$^2$~\cite{yoo2019photorealistic} introduced wavelet-corrected transfers that better preserve structural integrity and local feature statistics. DeepFaceEditing~\cite{chenDeepFaceEditing2021} further refines this approach by using local disentanglement and global fusion to more effectively separate and combine geometric and stylistic elements. BlendGAN~\cite{liu2021blendgan} adopts a self-supervised method, developing a style encoder that integrates a weighted blending module for seamless style integration. TargetCLIP~\cite{chefer2021targetclip} uses the StyleGAN2 latent space to identify desired editing direction that align with reference images, optimizing the CLIP similarity with the target. NeRFFaceEditing~\cite{NerfFaceEditing} utilizes appearance and geometry decoders in a tri-plane-based neural radiance field, using an AdaIN-based approach for enhanced decoupling of appearance and geometry. Different from these methods, our HyperGAN-CLIP model uses CLIP embeddings to dynamically control the modulation weights and decode the StyleGAN2 latent vectors, offering a more enhanced flexibility and precision in synthesis process. With the growing interest in diffusion models, there have been efforts to guide the denoising diffusion process using reference images as well. For example, diffusion frameworks in~\cite{balaji2022eDiff-I,bansal2024universal} allow image generation to be steered by the style of a reference image, while the content is specified by a text prompt. MimicBrush~\cite{Chen2024MimicBrush} builds on these works by enabling local semantic edits on input images using a reference image. This is achieved by automatically extracting the semantic correspondence between the input and reference images.

\begin{figure*}[!t]
  \centering
  \includegraphics[width=0.935\linewidth]{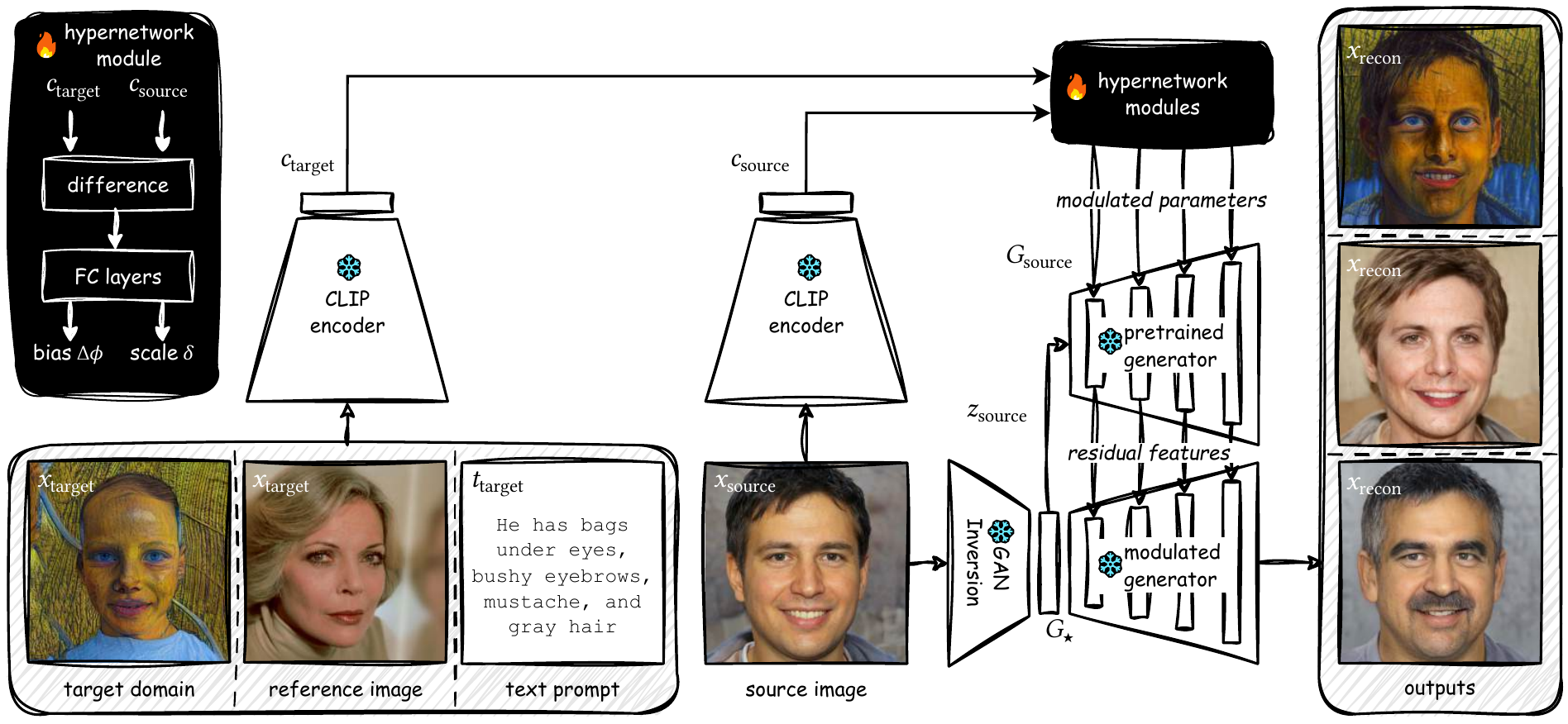}
  \caption{\textbf{Overview of HyperGAN-CLIP}. This framework employs hypernetwork modules to adjust StyleGAN generator weights based on images or text prompts. These inputs facilitate domain adaptation, attribute transfer, or image editing. The modulated weights blend with original features to produce images that align with specified domains or tasks like reference-guided synthesis and text-guided manipulation, while maintaining source integrity.}
  \label{fig:overview}
\end{figure*}

\subsection{Text-Guided Image Manipulation}
Text-guided image manipulation modifies images based on textual descriptions while preserving their structure and incorporating the specified attributes. Recent studies leverage CLIP~\cite{radford2021learning}, which provides a shared latent space for images and text, enabling precise text-driven editing. StyleCLIP-LO~\cite{Patashnik_2021_ICCV} optimizes latent codes to generate target images aligned with textual prompts. StyleCLIP-LM~\cite{Patashnik_2021_ICCV} predicts residual latent codes based on the CLIP similarity of attributes and output images. StyleCLIP-GD~\cite{Patashnik_2021_ICCV} maps text prompts to global directions in the original StyleGAN  space, while StyleMC~\cite{kocasari2021} explores global directions within StyleGAN2's lower dimensional $\mathcal{S}$ space to enhance this alignment. HairCLIP~\cite{wei2022hairclip} modulates latent codes for specific style attributes like hair color, using text for fine-grained control, optimizing similarity in the CLIP space. DeltaEdit~\cite{Lyu_2023_CVPR} trains latent mappers solely on images using semantically aligned $\Delta$-CLIP space, enabling manipulations guided by reference textual descriptions or images. CLIPInverter~\cite{CLIPInverter} conditions the inversion stage on textual descriptions, obtaining manipulation directions as residual latent codes through a CLIP-guided adapter module. In diffusion-based synthesis methods, DiffusionCLIP~\cite{Kim_2022_CVPR} modifies input images by first converting them to noise through forward diffusion and then guiding the reverse diffusion process using CLIP similarity to obtain the final image. Plug-and-play~\cite{pnpDiffusion2022} enhances image synthesis by injecting image feature maps from a latent diffusion model into the denoising process guided by textual descriptions. Pix2Pix-Zero~\cite{Parmar2023Pix2PixZero} maintains the structure of the original image with cross-attention guidance and applies targeted edits using an edit-direction embedding to modify specific objects. InstructPix2Pix~\cite{Brooks2023InstructPix2Pix} and MagicBrush~\cite{Zhang2023Magicbrush} enable semantic image editing based on user-provided textual instructions. ZONE~\cite{Li2024Zone} extends these approaches to zero-shot local image editing, utilizing the localization capabilities within pre-trained instruction-guided diffusion models.
%Unlike these works, except DeltaEdit, our approach utilizes the $\Delta$-CLIP space to guide a hypernetwork for zero-shot prediction of modulation weights, achieving efficient and accurate text-driven edits without relying on textual data during training.

\subsection{Hypernetworks}
Hypernetworks~\cite{ha2016hypernetworks} are neural networks designed to predict or modulate the weights of another network, known as the primary network. This ability enhances the flexibility and generalizability of models. %such as 3D modeling~\cite{littwin2019deep,spurek2021hyperpocket}, semantic segmentation~\cite{nirkin2021hyperseg}, and continual learning~\cite{ohs2019hypercl}. 
For instance, HyperInverter~\cite{dinh2021hyperinverter} employs hypernetworks to adjust encoder parameters, while HyperStyle~\cite{alaluf2021hyperstyle} uses them to adapt the StyleGAN generator, improving representation of out-of-domain images. DynaGAN~\cite{Kim2022DynaGAN} and HyperDomainNet~\cite{alanov2023hyperdomainnet} use hypernetworks for dynamic weight modulation in few-shot domain adaptation. Building on these, our method enhances StyleGAN’s adaptability by integrating hypernetworks with CLIP embeddings to modulate weights according to different modalities, letting our framework be used for both domain adaptation, reference-guided image synthesis and text-guided image manipulation.
\section{Approach}
\label{ssec:overview}
HyperGAN-CLIP represents a unified architecture built upon StyleGAN2~\cite{Karras_2020_CVPR}, designed to address a wide range of generative tasks such as domain adaptation, reference-guided image synthesis, and text-guided image manipulation. In Sec.~\ref{ssec:unified-framework}, we introduce the core components of HyperGAN-CLIP. Then, in Sec.~\ref{ssec:training}, we describe the training procedures employed to deploy HyperGAN-CLIP across the various generative and editing tasks. 

\subsection{HyperGAN-CLIP}
\label{ssec:unified-framework}
As shown in Fig.~\ref{fig:overview}, our HyperGAN-CLIP framework dynamically adjusts the weights of a StyleGAN2 generator pre-trained on a source domain using input images or text prompts. These versatile inputs can represent a target domain for adaptation, serve as an in-domain reference for attribute transfer, or function as a textual description for editing. This flexibility allows our framework to generate images that not only align with target domain characteristics but also support both reference-guided image synthesis and text-guided image manipulation, all while preserving the source domain's integrity.

At the core of HyperGAN-CLIP is a unified adaptation strategy that employs a single architecture to handle various generative tasks dynamically. This strategy centers around a hypernetwork module that interacts with each layer of a pre-trained StyleGAN generator to produce task-specific adaptations. However, rather than directly updating the original generator network, our approach involves updating the weights of a duplicated generator network. This network generates the missing features based on the provided CLIP~\cite{radford2021learning} embeddings of the conditioning inputs. These features are then integrated into the original, frozen generator network via a residual feature injection module, ensuring the preservation of the source domain's integrity.

More formally, the final features of a layer $i$, denoted by $F_i^{'}$, are estimated by injecting the scaled down modulated features $F_i^{*}$ into the original features $F_i$, as given below:
\begin{equation}
     F_i^{'} = F_i + \eta \cdot F_i^{*} \;, 
    \label{injection}
\end{equation}
where $\eta$ is the scaling parameter. By this way, the final features remain close to the original distribution at the beginning of the training process. The original intermediate features, $F_i$, are derived from the preceding layer's output $F_{i-1}^{'}$ using:
\begin{equation}
     F_i = F_{i-1}^{'} \circledast \theta_i + b_i \;,
    \label{conv_orig}
\end{equation}
with $\theta_i$ and $b_i$ respectively representing the layer weights and the layer bias of the pre-trained StyleGAN. Meanwhile, the modulated features, $F_i^{*}$, are computed using the weights $\theta_i^{*}$ modulated by the proposed CLIP-conditioned hypernetwork module as follows:
\begin{equation}
     F_i^{*} = F_{i-1} \circledast \theta_i^{*} + b_i \;,
    \label{conv}
\end{equation}
where the modulated weights, $\theta_i^{*}$ are defined as
\begin{equation}
     \theta_i^{*} =  \delta_i \cdot f(\phi_i + \Delta\phi_i, s_i)\;.
    \label{mod}
\end{equation}
Here, $f$ represents the composite function of cascaded modulation and demodulation operations, $s_i$ is the style vector transformed from the latent code $w$ of the source image, and $\phi_i$ denotes the convolutional weights of the pre-trained generator at layer $i$. Notably, the modulation parameters $\Delta\phi_i$ and $\delta_i$, the task-specific weight bias and the channel-wise scale parameter, are dynamically predicted by our proposed CLIP-conditioned hypernetwork module $H_i(\cdot)$, as:
\begin{equation}
     \Delta\phi_i, \delta_i = H_i(\Delta c)\;,
    \label{hyper}
\end{equation}
where $\Delta c$ is the $\Delta$-CLIP embedding~\cite{Lyu_2023_CVPR} representing the difference between the CLIP embedding of the conditioning input (an image or a text prompt) and the CLIP embedding of the source image. Each hypernetwork module is composed of two individual fully-connected layers that generate affine transformation parameters for each convolution layer, one for the weight bias matrix $\Delta{\phi_i}$ and the other for the weight scaling parameter $\delta_i$, respectively. Hence, the number of parameters introduced by the hypernetwork module depends on the length of $\Delta$-CLIP embeddings and the size of the corresponding convolutional layer, and often very less compared to the base generator network.
%The hypernetwork module computes these modulation parameters via fully-connected layers that perform affine transformations on $\Delta c$.%, enabling precise modulation control during the adaptation process.

Previous studies have shown that CLIP embeddings are effective at capturing the stylistic elements of reference images~\cite{balaji2022eDiff-I,bansal2024universal}. Utilizing $\Delta$-CLIP embeddings allows our model to focus solely on the attributes absent in the source domain, thereby eliminating any redundant information. This approach centers the input embeddings to the hypernetwork around zero, simplifying the training process. Moreover, our findings suggest that using raw CLIP embeddings directly can significantly change the identity and noticeably degrade image quality. A detailed analysis is given in the Supplementary Material. Another key outcome of using CLIP embeddings is that it allows for adapting the pre-trained generator to multiple domains with just a single network model. 

\subsection{Training HyperGAN-CLIP}
\label{ssec:training}
Consider $x$ as a synthetic image generated from noise or a natural image from the source domain $\mathcal{D}_{\text{source}}$. In the context of StyleGAN's architecture, $x$ is produced by the mapping $x = G_{\text{source}}(z)$, where $z$ is a latent vector either sampled from a noise distribution or derived using a GAN inversion technique. HyperGAN-CLIP is designed to adapt the pre-trained generator $G_{\text{source}}$ into a modulated generator $G_{\star}$. This adaptation enables $G_{\star}$ to handle multiple tasks: multiple domain adaptation, reference-guided image synthesis, and text-guided image manipulation. It accomplishes this by leveraging additional inputs, which may be specific images or text prompts, to customize the generator’s output to the requirements of these varied applications.
We train our HyperGAN-CLIP framework by minimizing a multi-task loss $\mathcal{L}$, defined as: 
\begin{eqnarray}
\mathcal{L} = \lambda_1\mathcal{L}_\text{CLIP}+\lambda_2\mathcal{L}_\text{CLIP-Across}+\lambda_3\mathcal{L}_\text{CLIP-Within}+\lambda_4\mathcal{L}_\text{cGAN}\nonumber \\
+\lambda_5\mathcal{L}_\text{Contrastive}
+\lambda_6\mathcal{L}_\text{ID}
+\lambda_7\mathcal{L}_\text{L2}+\lambda_8\mathcal{L}_\text{LPIPS}
\end{eqnarray}
where $\lambda_*$ depicts the corresponding regularization coefficients. %We next detail each of these terms. 

\subsubsection{CLIP-based Losses} 
For domain adaptation, the core objective is to align the semantics of the adapted domain images with those of a target domain image \(x_\text{target}\). We define \(z_\text{source}\) as the latent code corresponding to \(x_\text{target}\) inverted to the source domain, where it generates \(x_\text{fixed}\), the source domain equivalent of \(x_\text{target}\). The adapted generator aims to use the same \(z_\text{source}\) to produce an adapted image \(x_\text{recon}\). Leveraging the CLIP embeddings of the target images, we enforce semantic consistency through the CLIP similarity loss:
\begin{equation}
\mathcal{L}_\text{CLIP} = 1-\langle c_\text{recon}, c_\text{target}\rangle \;,
\end{equation}
where \(c_\text{target}\) and \(c_\text{recon}\) represent the CLIP embeddings of \(x_\text{target}\) and \(x_\text{recon}\), respectively, and $\langle \cdot, \cdot \rangle$ denotes the cosine similarity.

Global CLIP losses can lead to mode collapse and content loss~\cite{gal2021stylegannada}. Hence, as explored in~\cite{zhu2021mind}, we additionally adopt the following directional CLIP losses that measure the semantic shift within and across domains in CLIP space:
%to enhance diversity while preserving content:
\begin{eqnarray}
    \mathcal{L}_{\text{CLIP-Across}}=1-\langle \Delta c_\text{sample}, \Delta c_\text{fixed} \rangle\;, \\
    \mathcal{L}_{\text{CLIP-Within }}=1-\langle \Delta c_\text{source}, \Delta c_\text{target} \rangle \;.
\end{eqnarray}

\noindent To compute these losses, we begin by generating an image \( x_{\text{sample}} \) using the frozen generator \( G_{\text{source}} \) from a randomly sampled latent code. This image is then adapted to the target domain using \( G_{\star} \), resulting in \( x_{\text{trained}} \). Semantically, we anticipate that the differences between the source and target domains, captured by the \( \Delta \)-CLIP embeddings \( \Delta c_{\text{sample}} = \text{CLIP}(x_{\text{trained}}) - \text{CLIP}(x_{\text{sample}}) \) and \( \Delta c_{\text{fixed}} = \text{CLIP}(x_{\text{target}}) - \text{CLIP}(x_{\text{fixed}}) \), should align as they represent the transformation induced by domain adaptation. 

Additionally, to ensure the adaptation preserves essential semantic features across the transformation, the differences between source and adapted images, as measured by \( \Delta c_{\text{source}} = \text{CLIP}(x_{\text{fixed}}) - \text{CLIP}(x_{\text{sample}}) \) and \( \Delta c_{\text{target}} = \text{CLIP}(x_{\text{target}}) - \text{CLIP}(x_{\text{trained}}) \)), should also align.

For reference-guided image synthesis, HyperGAN-CLIP utilizes a refined methodology with in-domain data, adjusting StyleGAN's weights to faithfully replicate the style of target images. By leveraging pairs of source and target images from the source dataset, we effectively cover a broad distribution of CLIP embeddings, ensuring robust alignment between the CLIP space and StyleGAN image space. Specifically, we redefine $\mathcal{L}_{\text{CLIP-Across}}$ using the average StyleGAN image as the anchor image $x_\text{fixed}$, departing from the use of inverted target images typical in domain adaptation. During training, $x_\text{target}$ and $x_\text{sample}$ are randomly sampled. Furthermore, for $\mathcal{L}_{\text{CLIP-Within}}$, we substitute $x_\text{target}$ with $x_\text{recon}$ to enhance identity and content preservation. Please refer to the Supplementary Material for the graphical illustrations of these directional losses.

Notably, HyperGAN-CLIP trained for reference-guided image synthesis is also capable of performing text-guided image editing by using the \(\Delta\)-CLIP embedding \(\Delta c_\text{text} = \text{CLIP}(t_\text{target}) - \text{CLIP}(t_\text{source})\) to modulate the generator weights, with $t_\text{target}$ representing the input text prompt and $t_\text{source}$ denoting any text matching the source image. In our experiments, we use a generic prompt like ``\textit{face}’’ for $t_\text{source}$, but it can be replaced with a more fine-grained one. 

\subsubsection{CLIP-conditioned discriminator loss}
To preserve sample quality during domain adaptation, we introduce an adversarial loss \( \mathcal{L}_\text{cGAN} \) with a discriminator conditioned on CLIP embeddings. This discriminator, modeled after \cite{kumari2022ensembling,kang2023scaleup}, uses a frozen CLIP vision transformer backbone and only trains the outermost head layers. It dynamically measures the difference between source and target domain distributions. To deal with the data scarcity (we only have a single image per each target domain), we use differentiable augmentation \cite{zhao2020differentiable}. The conditioning of the discriminator on CLIP embeddings, implemented using a projection discriminator~\cite{miyato2018cgans}, ensures that the generated images align with the target domain characteristics and accelerates training convergence and prevents mode collapse.

\subsubsection{Contrastive Adaptation Loss}
To ensure that images generated from a target domain distinctly differ from those of other domains, we employ an adaptation loss \( \mathcal{L}_\text{Contrastive} \) encouraging the network to learn domain-specific transformations. Inspired by \cite{Kim2022DynaGAN}, this contrastive loss enhances similarity relationships, ensuring positive pairs (same domain) show higher similarity, while negative pairs (different domains) show less. Formally, it is given as:
%This loss, inspired by \cite{Kim2022DynaGAN}, uses contrastive learning to improve similarity relationships. Positive pairs (same domain) should exhibit higher similarity, while negative pairs (different domains) should show less. Formally, the loss is defined as:
\begin{equation}
    \mathcal{L}_\text{Contrastive} = -\log \frac{\exp(l_\text{pos})}{\exp(l_\text{pos}) + \Sigma_j \mathbf{1}_{[j \neq k]} \exp(l_\text{neg}^j)}
\end{equation}
with \( l_\text{pos} \), \( l^j_\text{neg} \) representing the cosine similarities of positive and negative pairs, respectively:
\begin{eqnarray}
    l_\text{pos} = \left\langle \text{CLIP}(x_\text{target}^k), \text{CLIP}(x_\text{recon}^k) \right\rangle \\
    l_\text{neg}^j = \left\langle \text{CLIP}(\text{Aug}(x_\text{target}^j)), \text{CLIP}(x_\text{recon}^k) \right\rangle
\end{eqnarray}
where \(\text{Aug}(\cdot)\) applies horizontal-flip and color-jitter augmentations to enhance training stability \cite{FuseDream2021}. This loss is calculated over a minibatch of 4 target domains for diverse domain learning.

\subsubsection{Identity Loss}
To preserve source identity when adapting to a target domain, we implement an identity similarity loss designed to maximize the cosine similarity between the image features from the source and target domains:
\begin{equation}
    \mathcal{L}_{\text{ID}} = 1 - \langle R(x_\text{sample}), R(x_\text{trained}) \rangle\,,
\end{equation}
where \( R(\cdot) \) extracts deep features using the ArcFace model \cite{Deng_2022}, specifically trained for face recognition. 

\subsubsection{Perceptual and Reconstruction Losses}
To complement the CLIP loss \( \mathcal{L}_{\text{CLIP}} \), we align \( x_{\text{recon}} \) with \( x_{\text{target}} \) using the L2 and LPIPS losses:
\begin{eqnarray}
    \mathcal{L}_{\text{L2}} = \|x_{\text{target}} - x_{\text{recon}}\|_2\\
    \mathcal{L}_{\text{LPIPS}} = \|F(x_{\text{target}}) - F(x_{\text{recon}})\|_2
\end{eqnarray}
where \( F(\cdot) \) represents AlexNet~\cite{Krizhevsky_AlexNet12} features.

\section{Experiments}

\subsection{Training and Implementation Details} 
We use the Adam optimizer with $\beta_{1}$ = 0.0 and $\beta_{2}$ = 0.99. We set the learning rate to 0.002 and the batch size to 4. For CLIP based losses, we use ViT-B/16 and ViT-B/32 CLIP encoder models and add their results as done in MTG.  We use the ViT-B/16 CLIP encoder while modulating the generator. The scaling parameter for the modulated features is set as $\eta = 0.1$ to prevent a large shift in feature distribution of the pretrained generator, ensuring stable training from the start. We empirically set the weights for the individual loss terms as
$\lambda_1=30$, $\lambda_2=1.5$, $\lambda_3=0.5$, $\lambda_4=0.2$, $\lambda_5=1.0$, $\lambda_6=3.0$, $\lambda_7=8.0$, and $\lambda_8=12.0$.
Each minibatch includes 4 randomly sampled target domain images $x_{\text{target}}$ and 4 source images $x_{\text{trained}}$.
%\revised{During training, we randomly sample 4 different target domain images $x_{target}$ for each minibatch. We also sample 4 random latents to generate images $x_{trained}$}. 
For domain adaptation and reference guided image synthesis, to find $x_\text{fixed}$ in the source domain corresponding to a target image, we use e4e inversion~\cite{Tov2021}. However, instead of using the inversion directly, we bring it closer to the mean latent by applying latent truncation. This prevents the inversion to lie in an out-of-distribution region and avoids $x_\text{fixed}$ and $x_\text{target}$ to be too close, and thus limiting meaningful editing directions.

\subsection{Domain Adaptation}
\label{sec:domain-adaptation}

\begin{figure*}[!t]
  \centering
\includegraphics[width=0.95\linewidth]{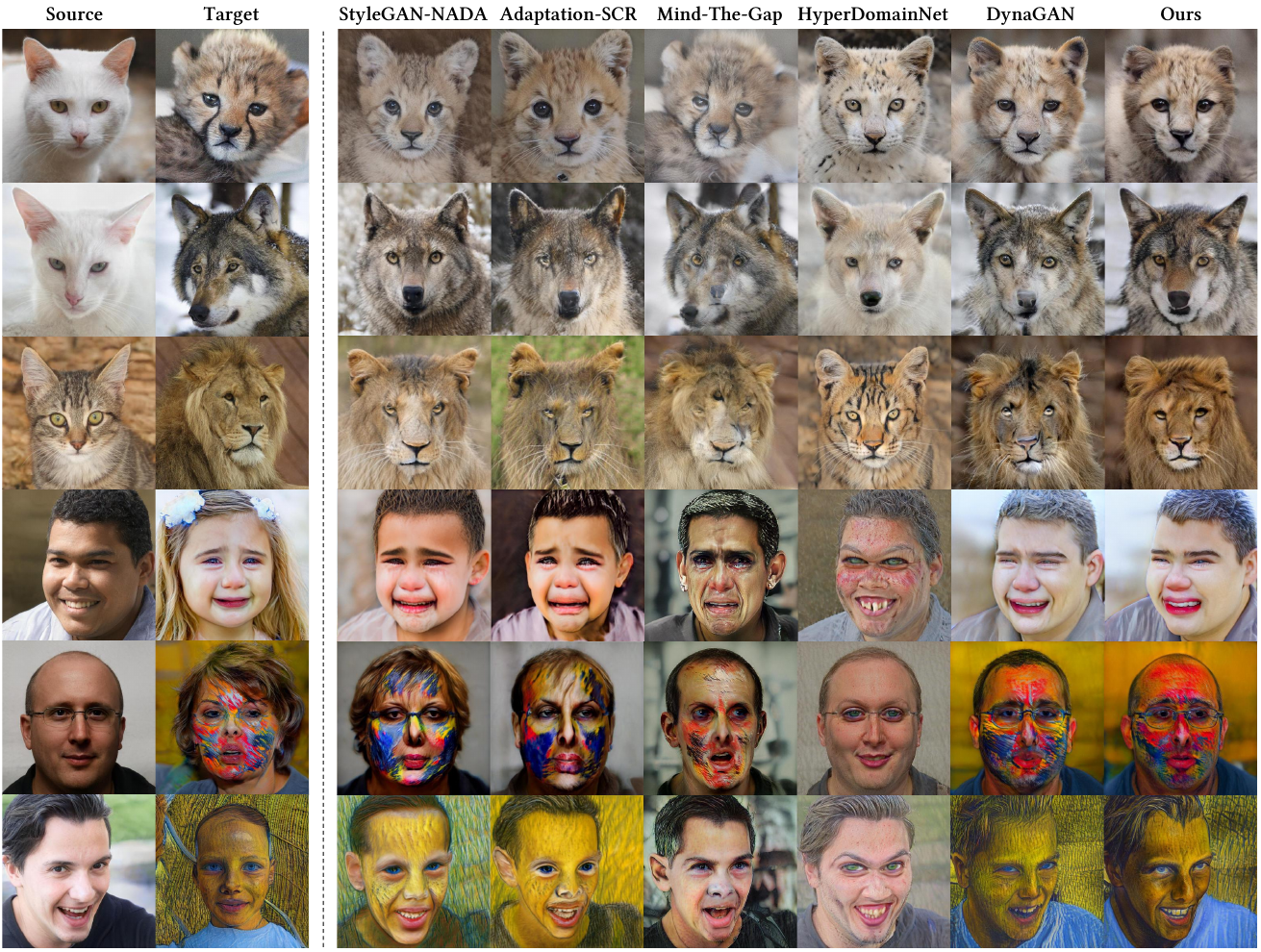}
  \caption{\textbf{Comparison against the state-of-the-art few-shot domain adaptation methods}. Our proposed HyperGAN-CLIP model outperforms competing methods in accurately capturing the visual characteristics of the target domains. 
  }
  \label{fig:dom_ada_qual}
\end{figure*}

We conduct two distinct experiments. First, we adapt a StyleGAN2 model, pre-trained on the FFHQ dataset~\cite{Karras_2019_CVPR}, to 101 new domains introduced in the expanded version of StyleGAN-NADA~\cite{gal2021stylegannada}. The training data was generated using the extended StyleGAN2 model provided by the authors of Domain Expansion~\cite{nitzan2023domain}\footnote{The NADA-expanded model used in our experiments is available at \url{https://github.com/adobe-research/domain-expansion/tree/main}.}. For each target domain, we sample a single image using the extended model, and use these sampled images to train our HyperGAN-CLIP model for multiple domain adaptation. Second, we use the AFHQ dataset to expand a StyleGAN2 model pre-trained on Cat images to 52 other animal domains (including 22 dog breeds and 30 wildlife animals represented by 7 cheetah, 6 tiger, 6 lion, 7 fox and 4 wolf images). For each target domain, we select a single image and use these samples to train HyperGAN-CLIP accordingly. We compare HyperGAN-CLIP to state-of-the-art GAN domain adaptation models, including Mind-the-GAP~\cite{zhu2021mind}, StyleGAN-NADA~\cite{gal2021stylegannada}, HyperDomainNet~\cite{alanov2023hyperdomainnet}, DynaGAN~\cite{Kim2022DynaGAN}, and Adaptation-SCR~\cite{Liu2023scr}. Each model is trained in the one-shot setting using the same training data. Notably, Mind-the-GAP, StyleGAN-NADA, and Adaptation-SCR require separate models for each target domain, whereas HyperDomainNet, DynaGAN, and HyperGAN-CLIP can model multiple domains with a single unified model. 
To quantitatively assess the quality and fidelity of the generated images, we adopt the widely used Fréchet Inception Distance (FID) score~\cite{FID} along with the Quality and Diversity metrics suggested in~\cite{alanov2023hyperdomainnet}. Details of these evaluation metrics are given in the Supplementary Material.

\begin{figure*}[!t]
    \centering
    \begin{subfigure}[t]{0.425\textwidth}
        \centering
        \begin{tabular}{ccc}
  $\;\;\quad$\small{Domain 1} & $\;\;\quad$\small{Domain 2} & $\;\;$\small{Hybrid}\\
  \multicolumn{3}{c}{\includegraphics[width=0.81\linewidth]{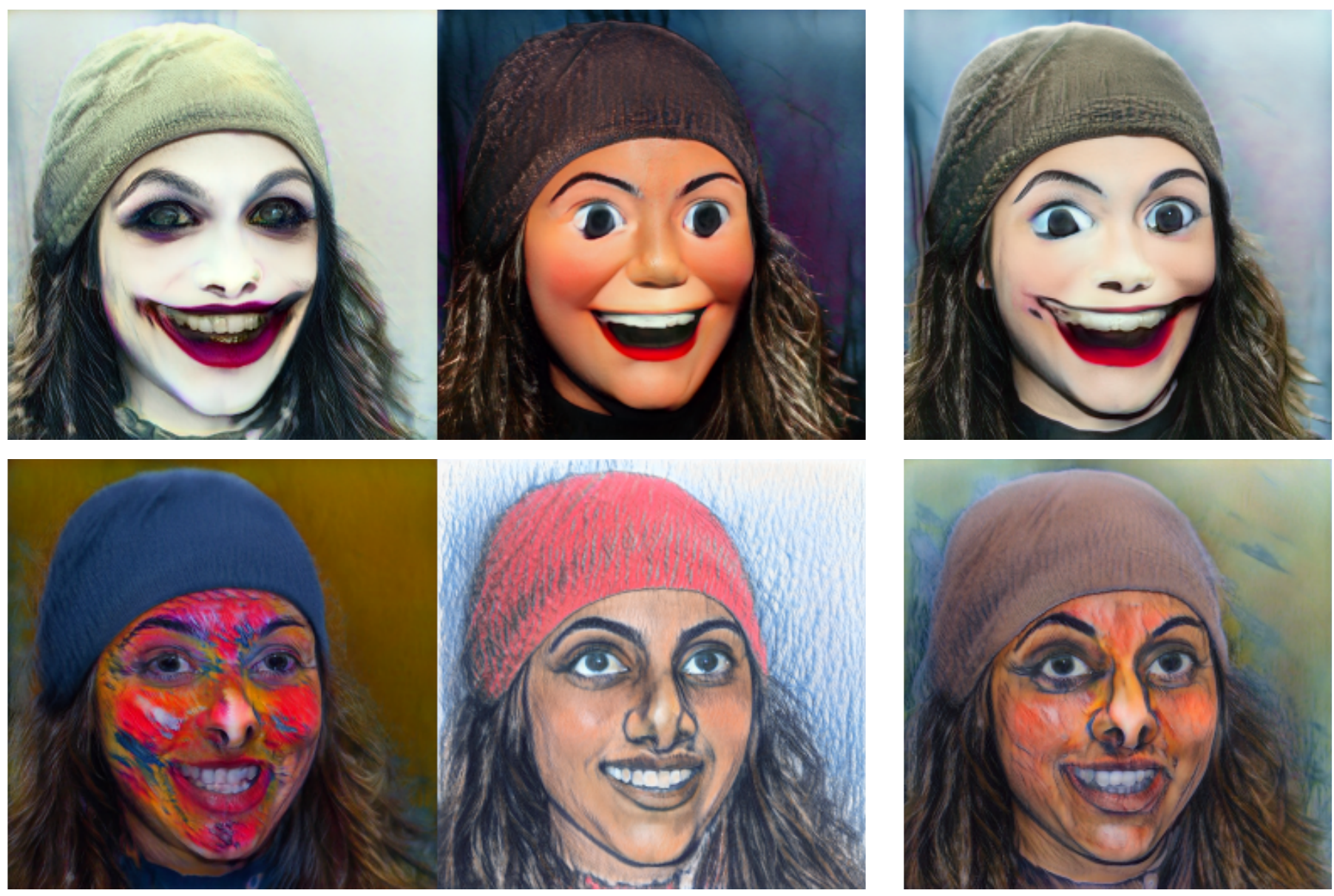}}
        \end{tabular}
        \caption{\textbf{Domain mixing.} Our approach can fuse multiple domains to create novel compositions. By averaging and re-scaling the CLIP embeddings of two target domains, we can generate images that blend characteristics from both.}
    \end{subfigure}%
    $\;\;$
    \begin{subfigure}[t]{0.555\textwidth}
        \centering
        \begin{tabular}{cccc}
  $\;\;\quad$\small{Adapted} & $\quad\qquad$\small{+Age} & $\quad\qquad$\small{+Smile} & $\quad$\small{+Pose}\\
  \multicolumn{4}{c}{\includegraphics[width=0.81\linewidth]{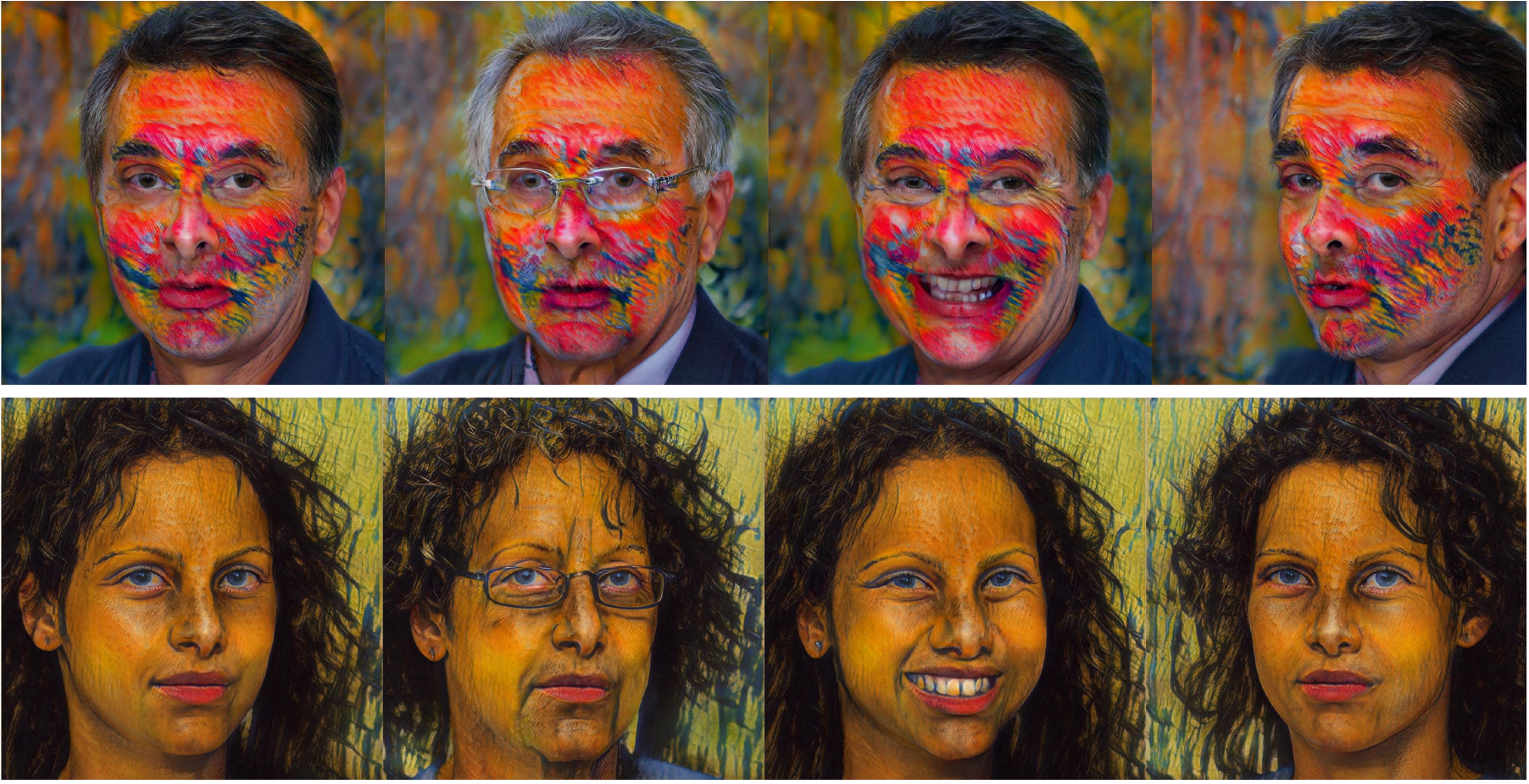}}
  \end{tabular}
        \caption{\textbf{Semantic editing in target domains.} Since latent mapper is kept intact, our approach allows for using existing latent space discovery methods to perform semantic edits. We manipulate two sample face images from adapted domains by playing with age, smile, and pose using InterfaceGAN~\cite{shen2020interfacegan}.}
    \end{subfigure}
    \caption{\textbf{Capabilities of HyperGAN-CLIP in blending domains and performing semantic edits within adapted domains.}}
    \label{fig:capabilities}
\end{figure*}

%\subsubsection{Results}
In Fig.~\ref{fig:dom_ada_qual}, we present sample images generated by the evaluated domain-adaptation techniques on the AFHQ and FFHQ datasets. Each sample includes the source image, the corresponding target domain training image and the synthesized outputs. Mind-the-Gap struggle to fully capture the visual characteristics of the target domains, often producing visually poor results. HyperDomainNet appears to have failed in learning very diverse domains, which leads to low-fidelity outcomes. While StyleGAN-NADA and Adaptation-SCR achieve better quality, they tend to slightly overfit to specific features of the representative target domain. DynaGAN shows improved performance over these models but sometimes generates unnatural and slightly distorted results, particularly in animal domains. It fails to fully reflect key features of the target domain, e.g., it does not generate desired small animal ears in the first row. Compared to DynaGAN, HyperGAN-CLIP better preserves source content. By leveraging CLIP-guided hypernetwork modules, it produces images with remarkable visual fidelity and effectively captures the essence of the target domains, as validated by the FID scores in Table \ref{tab:dom_ada_quantitative}. Moreover, the Diversity scores highlight that our approach demonstrates higher variability among the adapted images. Additional demonstrations of our model's ability to blend domains and perform semantic edits are given in Fig.\ref{fig:capabilities}. %We give additional comparisons, explore controllable image generation further, and present an ablation study in the Supplementary Material.
In the Supplementary Material, we provide additional comparisons, explore controllable image generation in  more detail, and present an ablation study. Moreover, we demonstrate that our approach can perform zero-shot domain adaptation relatively well on novel domains that are not semantically very different from the domains used during training.

\begin{table}[!t]
    \caption{\textbf{Quantitative results for multi domain adaptation.} HyperGAN-CLIP demonstrates strong performance in adapting characteristics of multiple target domains with a single model. The best and second best models are indicated in bold and underlined, respectively.
    }
    \resizebox{\linewidth}{!}{

    \begin{tabular}{lr@{\;\;}r@{\;\;}rr@{\;\;}r@{\;\;}r}
        \toprule
             & \multicolumn{3}{c}{AFHQ} & \multicolumn{3}{c}{FFHQ}\\          
             Method & FID$\downarrow$ & Qual$\uparrow$ & Div$\uparrow$ & FID$\downarrow$ & Qual$\uparrow$ & Div$\uparrow$\\          
        \midrule
Mind-The-Gap & 72.90 & \underline{0.93} & \underline{0.04} & 45.93 & 0.73 & 0.10 \\
StyleGAN-NADA & \underline{71.15} & \underline{0.93} & \underline{0.04} & 49.48 & 0.90 & 0.04 \\
Adaptation-SCR & \textbf{70.84} & 0.92 & 0.03 & 45.88 & 0.59 & 0.06\vspace{0.075cm}     \\
\hline \vspace{-0.3cm}\\
HyperDomainNet & 105.90 & 0.78 & \textbf{0.05} & 100.92 & 0.67 & 0.11\\
DynaGAN & 72.16 & \textbf{0.94} & 0.02 & \underline{28.94} & \textbf{0.83} & {0.14} \\ 
%Ours & \textbf{70.63} & \textbf{0.94} & 0.03 & 30.55 & \underline{0.82} & \underline{0.18} \\ 
Ours% (No trunc. + stylemix) 
& 71.93 & \textbf{0.94} & \underline{0.04} & \textbf{24.74} & \underline{0.81} & \textbf{0.16} \\ 
%Ours (No trunc. + no stylemix) & \textbf{69.45} & \textbf{0.94} & \underline{0.04} & 26.57 & \underline{0.82} & \textbf{0.20} \\ 
         \bottomrule
\end{tabular}

}
\label{tab:dom_ada_quantitative}
\end{table}

\subsection{Reference-Guided Image Synthesis}
\label{sec:reference-guided}

%\subsubsection{Experimental Setup}
In this experiment, our objective is to synthesize a new image that combines the identity of a source image with the style of a target image, as represented by its CLIP embedding. For quantitative analysis, we use the test set of the CelebA-HQ dataset~\cite{CelebAMask-HQ}, which comprises a total of 6000 diverse images, as the source and the target images. We assign a different target image to each source image by making sure that the same image is not used as source and target. We invert the source images to the latent space using an e4e encoder~\cite{Tov2021} pre-trained on the FFHQ dataset. The inverted latents are fed to our framework along with the CLIP embedding obtained from the target image to synthesize the final output. 
%\noindent\textbf{Competing Approaches.}
We compare HyperGAN-CLIP against % four state-of-the-art approaches. These approaches are 
BlendGAN~\cite{liu2021blendgan}, TargetCLIP-O~\cite{chefer2021targetclip}, TargetCLIP-E~\cite{chefer2021targetclip}, and MimicBrush~\cite{Chen2024MimicBrush}. While BlendGAN and TargetCLIP-E are encoder-based approaches, TargetCLIP-O employs a direct optimization scheme, and MimicBrush is a diffusion based approach (the whole image region is used as the input mask). Our approach, apart from these studies, is based on modulating the StyleGAN generator via CLIP-guided hypernetworks.

\begin{figure*}[!t]
  \centering  \includegraphics[width=0.755\linewidth]{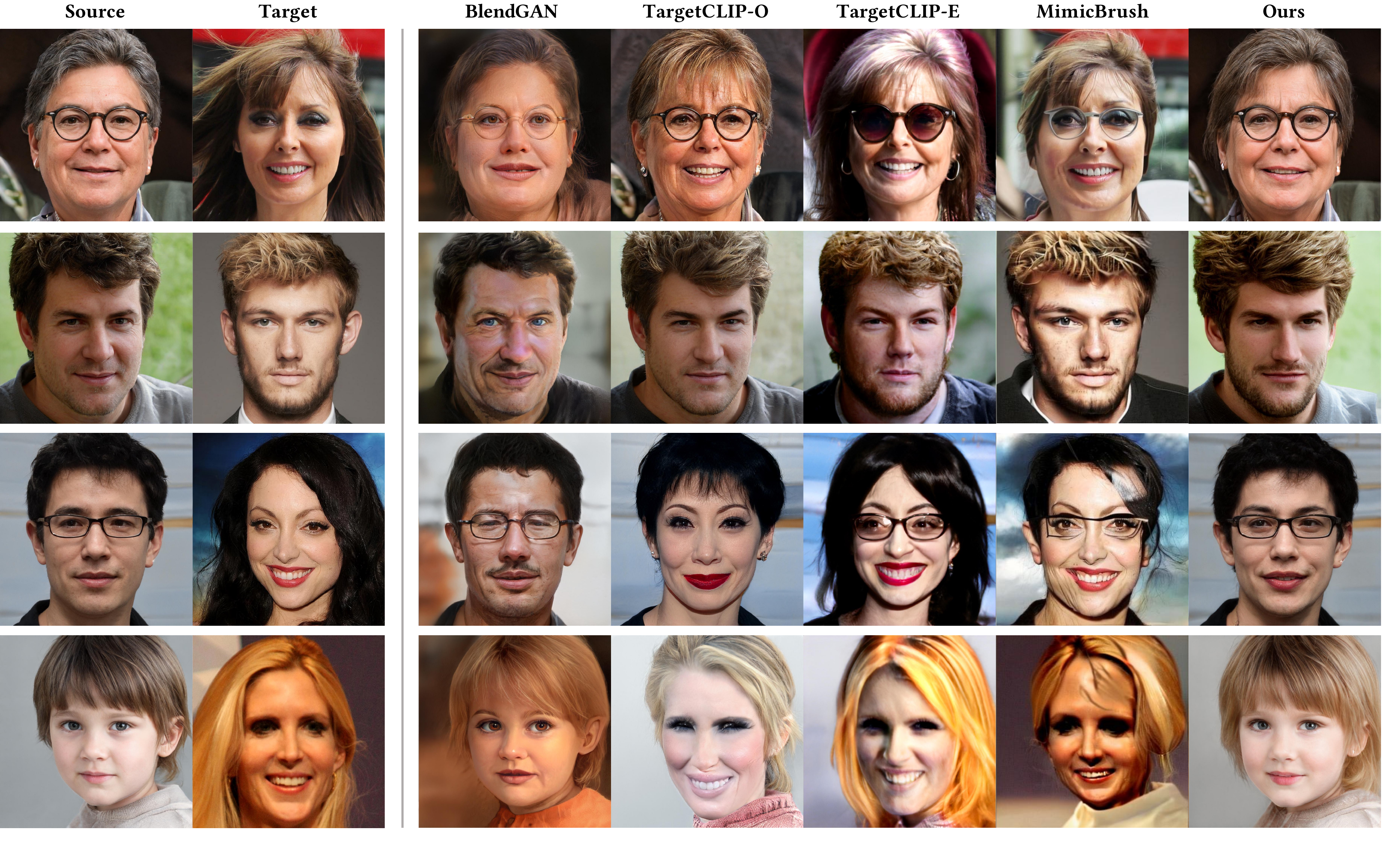}
  \caption{\textbf{Comparison with state-of-the-art reference-guided image synthesis approaches.} Our approach effectively transfers the style of the target image to the source image while effectively preserving identity compared to competing methods. 
  }
  \label{fig:image_based_qualitative}
\end{figure*}

%\subsubsection{Results}
In Fig.~\ref{fig:image_based_qualitative}, we present sample qualitative comparisons. %with existing methods. 
Sample source-target pairs show a diverse range of visual characteristics in terms of gender, age, hair color, ethnicity. BlendGAN tends to produce cartoon-like outputs that lack naturalness. Optimization-based TargetCLIP-O shows superior performance compared to its encoder-based counterpart TargetCLIP-E in maintaining identity while incorporating the desired style changes depicted in the target image. MimicBrush directly copies the target face onto the source pose, failing to transfer just the style and often resulting in unrealistic outputs. Notably, HyperGAN-CLIP gives superior performance in seamlessly transferring the attributes from the chosen target faces to the source faces while preserving identity to a greater extent than the competing methods. These results affirm the effectiveness of our approach in generating visually compelling outputs with enhanced fidelity and plausibility. Table~\ref{tab:image_based_quantitative} shows the quantitative results. Our method achieves competitive results in terms of FID, better than TargetCLIP-O, which performs latent optimization for each target. This highlights our method's ability to generate high-quality and faithful images. Moreover, our approach outperforms competing methods in preserving the identity of the source image, as indicated by the ID similarity scores. Additionally, our method excels in CLIP semantic similarity, affirming its capability to capture the semantics of the target image in the synthesized results. Overall, our approach strikes a favorable balance across multiple evaluation metrics, showing its effectiveness in photo-realistic image synthesis and preserving key visual attributes. 

\begin{figure}[!t]
  \centering
\includegraphics[width=0.95\linewidth]{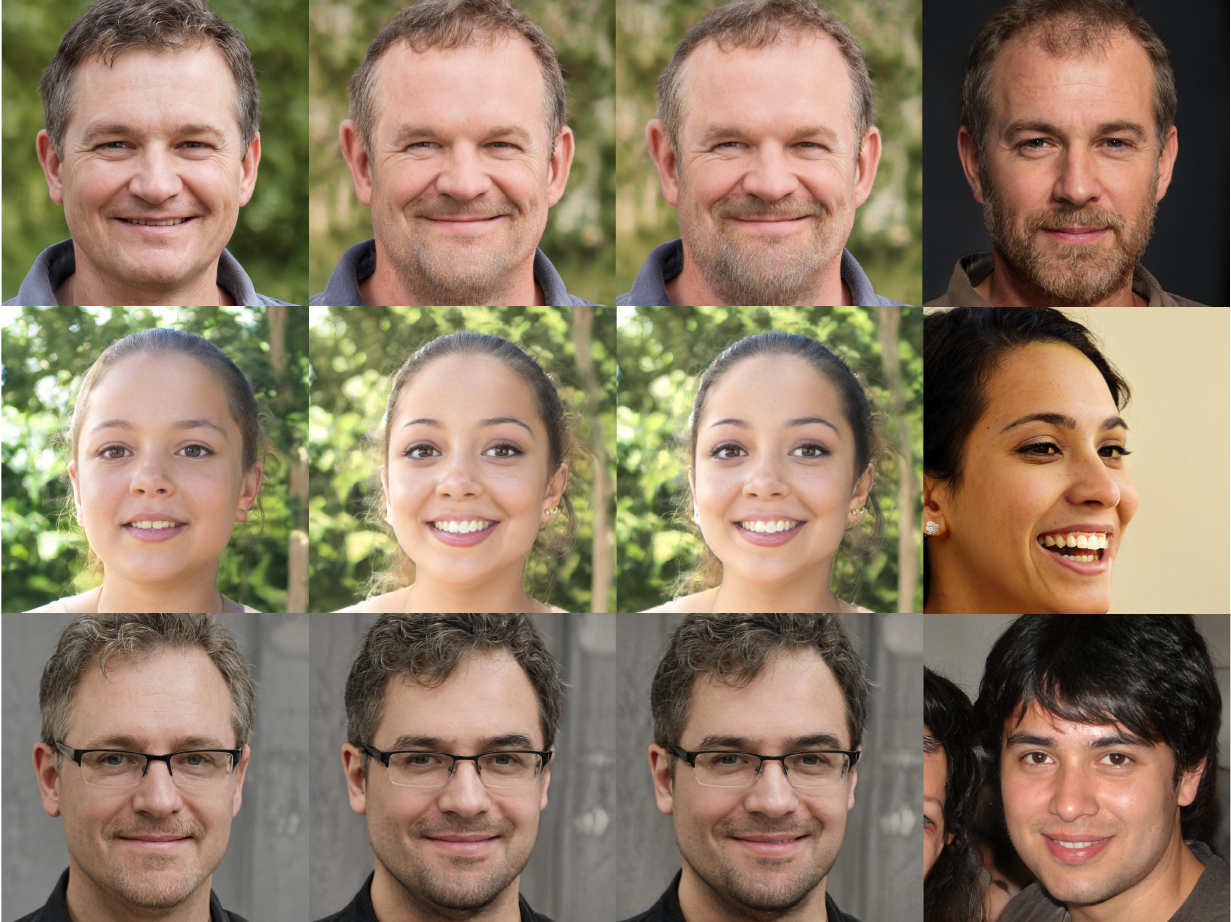}
  \caption{\textbf{Reference-guided image synthesis with mixed embeddings}. Each row shows the input image, the initial result with the CLIP image embedding, the refined result with a mixed embedding that incorporates the target attribute with $\alpha=0.5$, and the reference image, respectively. Target text attributes are \textit{``beard''} (top row), \textit{``black hair''} (middle row), and \textit{``smiling''} (bottom row). Incorporating mixed modality embeddings results in more accurate and detailed image modifications.} 
  \label{fig:mixed_embedding}
\end{figure}

One key limitation of both our proposed method and the competitive approaches is that, in some cases, they struggle to transfer fine attributes from reference images because their global image embeddings lack the specificity needed to capture these details. To address this issue, we explore a strategy that combines the CLIP embeddings of reference images with those of text prompts designed to capture specific target attributes. By leveraging CLIP's capability to encode both visual and textual data, we refine the reference image embedding by incrementally adding the embedding of the target attribute, modulated by an $\alpha$ parameter, following the formula $\text{CLIP}(x_\text{target})+\alpha\;\text{CLIP}(t_\text{target})$. As demonstrated in Fig.~\ref{fig:mixed_embedding}, this strategy enhances the editing process by allowing fine-tuned adjustments to specified attributes, resulting in more accurate and detailed image modifications based on the reference image.

\begin{table}[!t]
    \caption{\textbf{Quantitative results for reference-guided image synthesis.} HyperGAN-CLIP outperforms the existing models, generating high-quality images. It effectively preserves source identity while transferring the semantic details of the target images. The best and second-best models are highlighted in bold and underlined, respectively.}
    %\resizebox{0.98\linewidth}{!}{  
\begin{tabular}{l@{\;\;}r@{\;\;}r@{\;\;}r@{\;\;}r}
        \toprule
             Method & FID$\downarrow$ & ID (source)$\uparrow$ & ID (target)$\downarrow$ & CLIP Sim.$\uparrow$  \\
             
        \midrule
BlendGAN & 14.54 & 34.58±9.91 & \textbf{2.63±9.53} & 77.08±7.17 \\
TargetCLIP-O & \underline{11.26} & \underline{50.77±16.61} & 17.78±10.54 & 77.16±9.71  \\
TargetCLIP-E & 29.48 & 41.51±11.61 & 26.94±10.40 & 72.41±8.01 \\
MimicBrush & 37.06 & 11.19±10.43 & 65.91±14.65 & \underline{82.69±7.29} \\
Ours      & \textbf{8.73} & \textbf{78.73±6.01} & \underline{10.51±10.04} & \textbf{90.78±3.80} \\
         \bottomrule
\end{tabular}
%}
\label{tab:image_based_quantitative}
\end{table}

\begin{figure*}[!t]
  \centering
\includegraphics[width=0.99\linewidth]{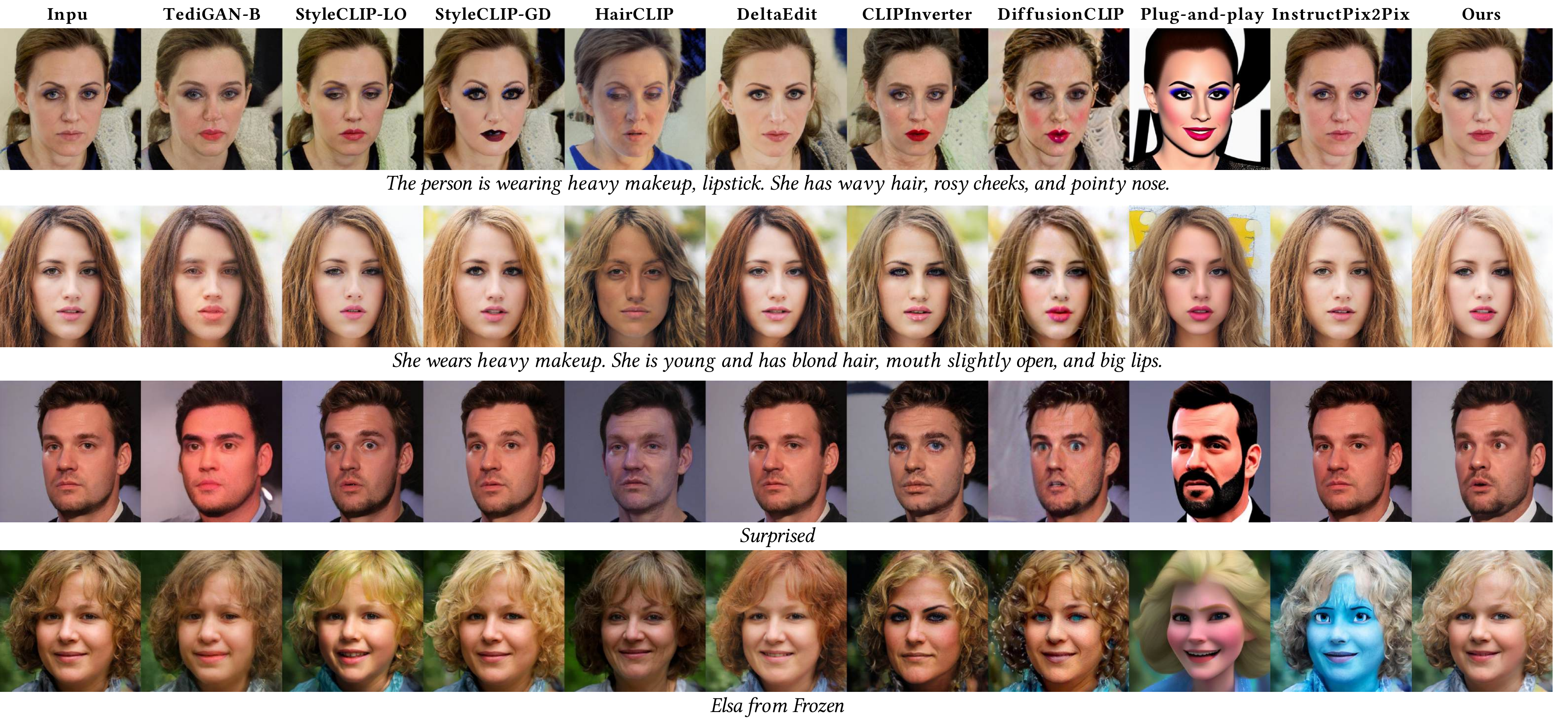}
\caption{\textbf{Comparisons with state-of-the-art text-guided image manipulation methods.} Our model shows remarkable versality in manipulating images across a diverse range of textual descriptions. The results vividly illustrate our model's ability to accurately apply changes based on target descriptions encompassing both single and multiple attributes. Compared to the competing approaches, our model preserves the identity of the input much better while successfully executing the desired manipulations. 
}
\label{fig:text_based_qualitative}
\end{figure*}

\subsection{Text-Guided Image Manipulation}
\label{sec:text-guided}
%\subsubsection{Experimental Setup} 
In this experiment, we show the versatility of our proposed framework by demonstrating its ability to manipulate input images based on target textual descriptions. For the quantitative analysis, we leverage the CelebA dataset's test set~\cite{liu2015faceattributes} along with its attribute annotations. We select attributes that are absent from the images and construct target descriptions that prompt the desired attribute manipulation. Leveraging a pre-trained e4e model~\cite{Tov2021}, we perform an image-to-latent-space inversion, generating latent representations of the input images. These inverted images serve as inputs to our framework. To condition the synthesis process, we utilize $\Delta$-CLIP embeddings, which capture the discrepancy between the CLIP embeddings of the target description and the input image.
%\noindent\textbf{Competing Approaches.}
We perform a comprehensive comparison of our method against several state-of-the-art text-guided image manipulation approaches. These include TediGAN-B~\cite{xia2021tedigan}, StyleCLIP-LO~\cite{Patashnik_2021_ICCV}, StyleCLIP-GD~\cite{Patashnik_2021_ICCV}, HairCLIP~\cite{wei2022hairclip}, DeltaEdit~\cite{Lyu_2023_CVPR}, and CLIPInverter~\cite{CLIPInverter} as representative GAN-based methods. Among these, DeltaEdit is the only model that utilizes text-free training like our method. Additionally, we also compare against diffusion-based approaches, namely DiffusionCLIP~\cite{Kim_2022_CVPR}, Plug-and-Play~\cite{pnpDiffusion2022}, and InstructPix2Pix~\cite{Brooks2023InstructPix2Pix}. Among these, the method most similar to ours is DeltaEdit in the sense that it is also solely trained on image data and does not utilize any text data during training. By evaluating our method against these diverse approaches, we provide a comprehensive analysis of its performance and highlight its distinct advantages in text-guided image manipulation.
To evaluate the approaches quantitatively, we employ Fréchet Inception Distance (FID) \cite{FID}, Attribute Manipulation Accuracy (AMA), and CLIP Manipulative Precision (CMP) following the methodology introduced by CLIPInverter~\cite{CLIPInverter}. Please refer to the supplementary material for more details on the evaluation metrics. 

%\subsubsection{Results}
%\canberk{Paragraph revised according to the new figure}
Fig.~\ref{fig:text_based_qualitative} presents text-guided image manipulation results of our proposed approach along with several competing methods across various textual descriptions. TediGAN-B and DeltaEdit struggle to effectively manipulate the images, often resulting in images similar to the input. While StyleCLIP-LO, StyleCLIP-GD and HairCLIP perform better, they still exhibit limitations when manipulating all specified attributes.%certain attributes (e.g., ``\textit{bangs}'' in row 1). 
 CLIPInverter performs well when explicit attribute manipulations are specified in the descriptions (first two rows), but it falls short when encountering novel descriptions unseen during its training, such as ``\textit{surprised}'' or ``\textit{Elsa from Frozen}''. DiffusionCLIP~\cite{Kim_2022_CVPR} generates images with noticeable artifacts, leading to poor output quality. While Plug-and-play~\cite{pnpDiffusion2022} successfully applies most manipulations, the resulting images often lack realism, appearing cartoonish and with unintended attribute modifications. In contrast, our model, even trained without any textual data, successfully applies single or multiple attribute changes while better preserving the identity of the input images compared to the competing approaches. 

\begin{table}[!t]
    \caption{\textbf{Quantitative results for text-guided image editing.} Even without explicit training on textual descriptions, HyperGAN-CLIP achieves results competitive with the state-of-the-art methods. The best and second best models are highlighted in bold and underlined, respectively.}
    \resizebox{0.8\linewidth}{!}{  
\begin{tabular}{lrrrr}
        \toprule
             & {FID$\downarrow$} & {CMP$\uparrow$} & {AMA$\uparrow$}& {AMA$\uparrow$}\\
             &  &  & (Sng.)& (Mult.)\\             
             %& & & (Single) & (Multiple) \\
        \midrule 
TediGAN-B & 55.424 & \textbf{0.285} & 11.286 & 1.142 \\
StyleCLIP-LO & 80.833                   & 0.210                  & 15.857                         & 3.429                              \\
StyleCLIP-GD & 82.393                   & 0.191                  & 33.143                            & 11.429                             \\
HairCLIP     & 93.523                   & 0.218                  & \underline{41.571}                            & 15.149                             \\
CLIPInverter         & 97.210                        & 0.221                       & \textbf{61.429}                                  & \textbf{41.714}   \\
DiffusionCLIP      & \textbf{29.280}                   & \underline{0.243}                  & 26.000                           & 4.857                              \\
Plug-and-play      & 68.287                   & 0.199                  & 27.429                           & 7.143  \\
InstructPix2Pix & \underline{47.531} & 0.173 & 40.571 & \underline{19.714} \vspace{0.075cm}\\
\hline \vspace{-0.3cm}\\
DeltaEdit      & 80.316                   & 0.171                  & 8.857                           & 0.571                              \\
Ours         & 87.851                        & 0.189                       & 25.143                                  & 10.000   \\
         \bottomrule
\end{tabular}
}
\label{tab:text_based_quantitative}
\end{table}

Table~\ref{tab:text_based_quantitative} presents the quantitative results. Here, we group our approach and DeltaEdit together to distinguish these works from the others which utilize additional text data during training. We evaluate manipulation accuracy and precision using AMA (Single) for single attribute changes and AMA (Multiple) for multiple attribute changes. Remarkably, our model achieves comparable or even better performance in manipulation accuracy and precision compared to leading text-guided image manipulation models, including StyleCLIP, and DiffusionCLIP. In terms of FID, the diffusion-based models, DiffusionCLIP and Plug-and-play, excel as compared to GAN-based approaches due to their high-quality generation capabilities. Even though we do not use textual data during training, our model finds a good balance between the metrics and consistently delivers competitive performance. It effectively handles descriptions involving multiple attribute changes. More importantly, as compared to DeltaEdit, the other text-guided image manipulation method with text-free training, our HyperGAN-CLIP gives much superior performance.

In the Supplementary Material, we provide further visual comparisons and example results on the CUB-Birds dataset for reference-guided image synthesis and text-guided image manipulation tasks. In addition to the quantitative analyses, we conducted a user study using Qualtrics with 16 participants to evaluate the performance of the models for all three tasks. We focused on methods that have similar characteristics to ours: all-in-one models for multiple domain adaptation and text-based editing methods with text-free training. In our human evaluation, we randomly generated 25 questions for each task and asked participants to rank the models based on their performance. The rankings showed that our HyperGAN-CLIP model, using a single unified framework, achieves highly competitive results, often outperforming or matching the existing models. For more details, please refer to the Supplementary Material.

\section{Conclusion}
We present HyperGAN-CLIP, a flexible framework  for addressing domain adaptation challenges in GANs, also supporting both reference-guided image synthesis and text-guided image manipulation. Our efficient hypernetwork modules adapt a pre-trained StyleGAN generator to handle both image and text inputs. By utilizing residual feature injection and a conditional discriminator, it preserves source identity and image diversity while effective transferring target domain characteristics to produce high-fidelity images. Extensive evaluations show that HyperGAN-CLIP outperforms existing domain adaptation methods, excels in text-guided editing, and competes strongly in reference-guided image synthesis. While our framework handles various tasks, some require distinct training processes. Future research could seamlessly incorporate a mixture-of-experts approach to train a single model equipped with routing mechanisms.

\begin{acks}
This work was supported by KUIS AI Fellowships to ABA, ACB and MBK, Cambridge Trust \& Computer Science Premium Scholarship to ACB, TUBA GEBIP 2018 Award to EE, BAGEP 2021 Award to AE, and an Adobe research gift. 
\end{acks}

\newpage
\bibliographystyle{ACM-Reference-Format}
\bibliography{sample-bibliography}

\end{document}

% Title portion
%\title{Text-Guided Image Manipulation Using GAN Inversion}
\title{Supplementary Material: HyperGAN-CLIP: A Unified Framework for Domain Adaptation, Image Synthesis and Manipulation}

\author{Abdul Basit Anees}
\email{abdulbasitanees98@gmail.com}
\orcid{0000-0003-1293-1796}
\affiliation{%
  \institution{Koç University}
  \country{Turkey}
}

\author{Ahmet Canberk Baykal}
\email{canberk.baykal1@gmail.com}
\orcid{0000-0002-0249-5858}
\affiliation{%
  \institution{University of Cambridge}
  \country{United Kingdom}
}

\author{Muhammed Burak Kizil}
\email{mkizil19@ku.edu.tr}
\orcid{0009-0008-9007-9280}
\affiliation{%
  \institution{Koç University}
  \country{Turkey}
}

\author{Duygu Ceylan}
\email{duygu.ceylan@gmail.com}
 \orcid{0000-0002-2307-9052}
\affiliation{%
  \institution{Adobe Research}
  \country{United Kingdom}
}

\author{Erkut Erdem}
\email{erkut@cs.hacettepe.edu.tr}
\orcid{0000-0002-6744-8614}
\affiliation{%
  \institution{Hacettepe University}
  \country{Turkey}
}

\author{Aykut Erdem}
\email{aerdem@ku.edu.tr}
\orcid{0000-0002-6280-8422}
\affiliation{%
  \institution{Koç University}
  \country{Turkey}
}

\renewcommand{\shortauthors}{Anees et al.}

\begin{CCSXML}
<ccs2012>
<concept>
<concept_id>10010147.10010371.10010382</concept_id>
<concept_desc>Computing methodologies~Image manipulation</concept_desc>
<concept_significance>500</concept_significance>
</concept>
<concept>
<concept_id>10010147.10010257.10010293.10010294</concept_id>
<concept_desc>Computing methodologies~Neural networks</concept_desc>
<concept_significance>500</concept_significance>
</concept>
</ccs2012>
\end{CCSXML}

\ccsdesc[500]{Computing methodologies~Image manipulation}
\ccsdesc[500]{Computing methodologies~Neural networks}

\maketitle

The purpose of this document is to provide extra material to complement the main paper. Section~\ref{sec:directional-clip} presents graphical illustration of the directional CLIP losses for a better understanding of these losses and their roles. Section~\ref{sec:eval_setup} provides details about the evaluation setups used in domain adaptation, reference-guided image synthesis, and text-guided image manipulation experiments. 
Section~\ref{sec:user_study} gives the details of the user study that was carried out to assess the model performances on 

Section~\ref{sec:controllable_generation} explores the controllable generation of images by manipulating the scaling of CLIP embeddings and style mixing. Section~\ref{sec:ablation} presents the results of our ablation study, highlighting the contribution of each component of our model to overall performance. Section~\ref{sec:delta_space} demonstrates  qualitative comparisons between using raw CLIP embeddings and $\Delta$-CLIP embeddings for text-guided editing. Section~\ref{sec:additional_results} presents additional visual comparisons between our proposed model and existing approaches in few-shot domain adaptation, reference-guided image synthesis, and text-guided image manipulation. Section~\ref{sec:limitations} and Section~\ref{sec:ethical_statement} discusses the limitations and ethical implications of our work, respectively.

\section{Directional CLIP losses}
\label{sec:directional-clip}

\begin{figure*}[!t]
    \centering
    \begin{subfigure}[t]{0.45\textwidth}
        \centering
    {\includegraphics[width=\linewidth]{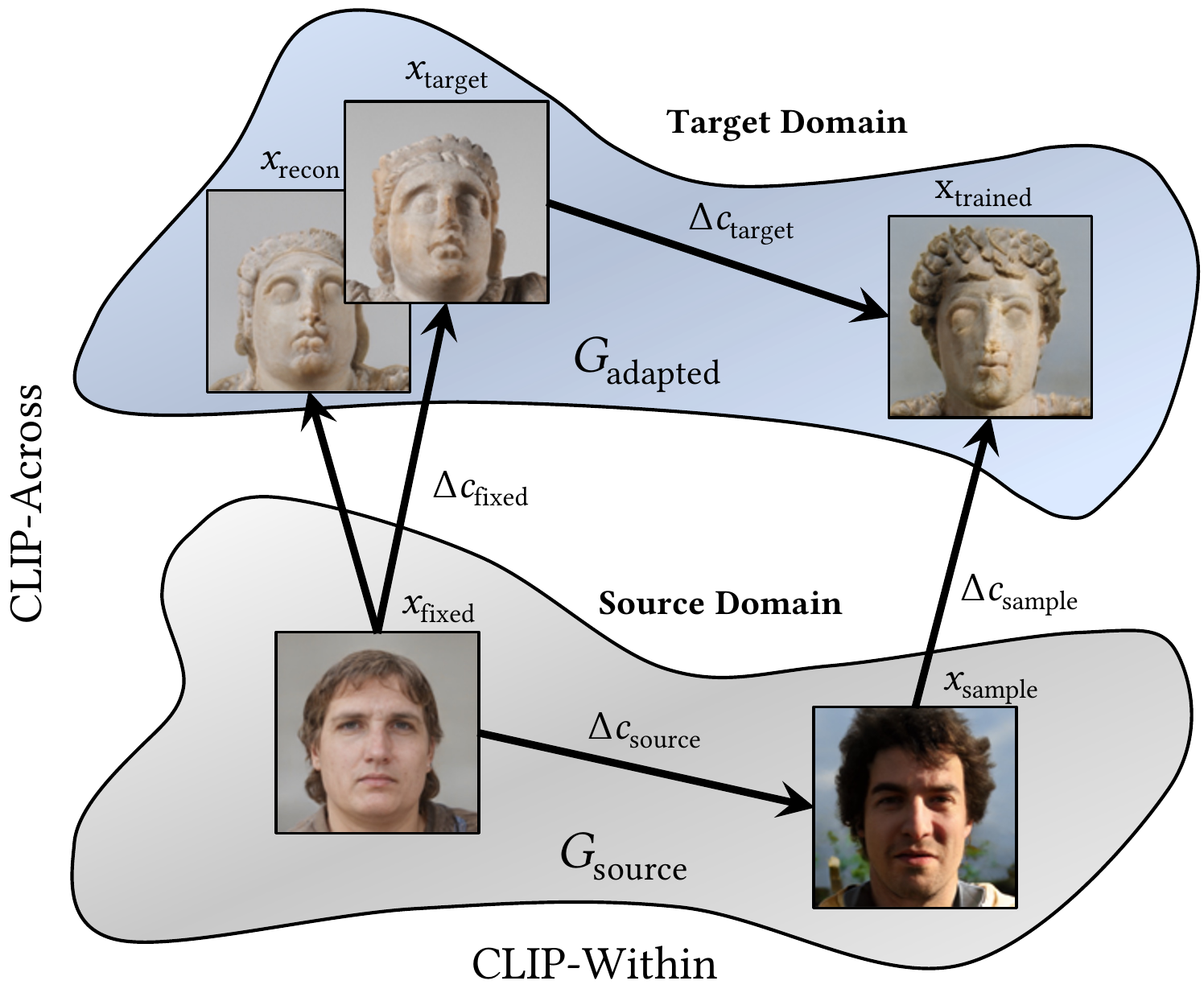}}
        \caption{\textbf{For domain adaptation.} We encode the images generated by the original and the modulated generators, representing the source and target domains, in the CLIP space. CLIP-Across loss, involving $\Delta c_\text{sample}$ and $\Delta c_\text{fixed}$, captures the differences between the source and target domains. On the other hand, CLIP-Within loss, computed using $\Delta c_\text{source}$ and $\Delta c_\text{target}$, preserves the semantic information that is unrelated to the domain gap.}
    \end{subfigure}%
    $\;\;$
    \begin{subfigure}[t]{0.45\textwidth}
        \centering
        {\includegraphics[width=\linewidth]{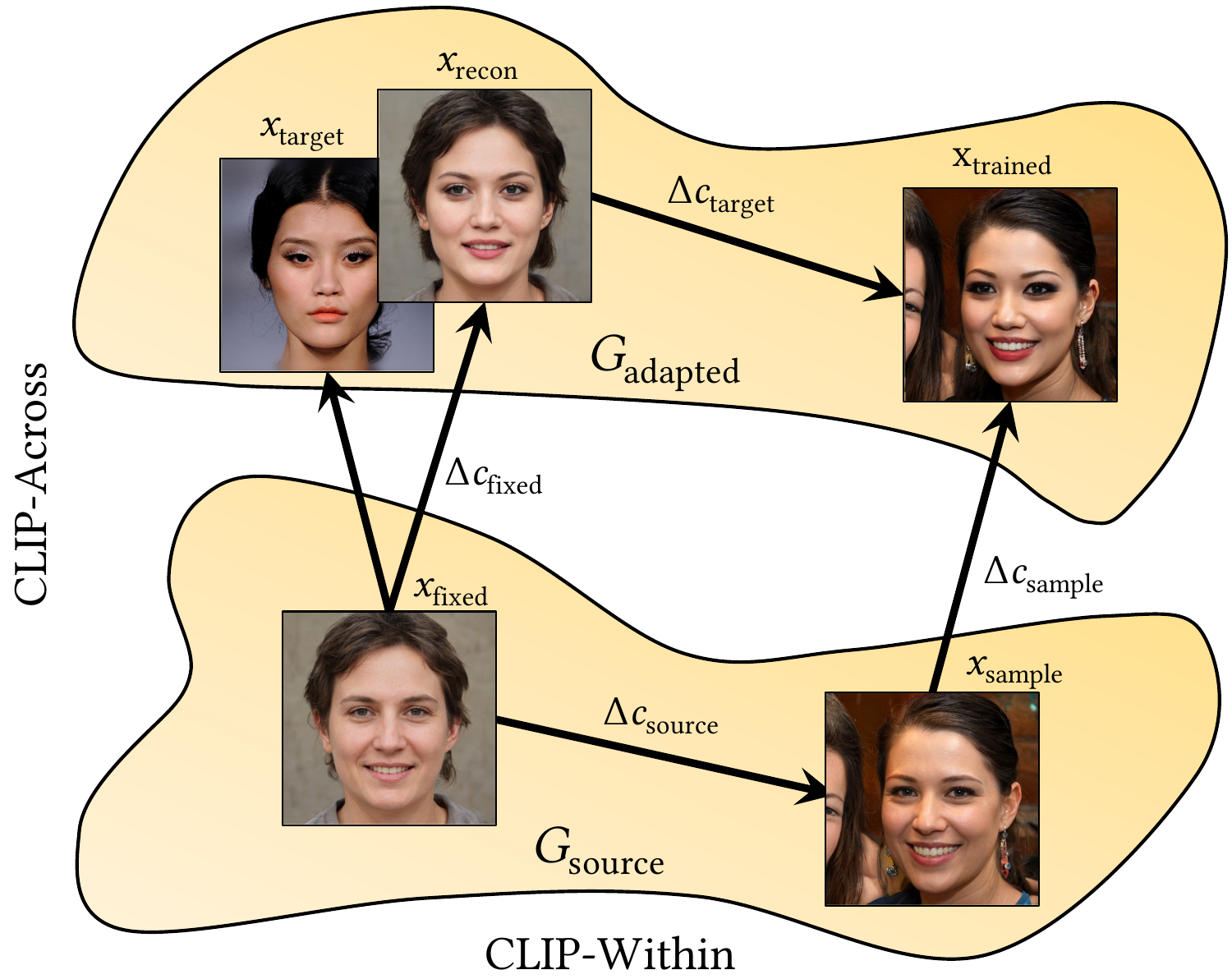}}
        \caption{\textbf{For reference-guided image synthesis.} In reference-guided image synthesis, source and target domains are the same, and thus it involves in-domain adaptation. CLIP-Across loss uses the mean StyleGAN image as the anchor image $x_\text{fixed}$. On the other hand, CLIP-Within loss utilizes the reconstructed image $x_\text{recon}$ to better preserve facial identity and image content.}
    \end{subfigure}
    \caption{\textbf{Visualization of the directional CLIP losses.} (a) for domain adaption. (b) for reference-guided image synthesis.}
    \label{fig:directional_losses}
\end{figure*}

Together, these losses ensure that crucial semantic information is preserved while capturing variations unique to each domain. Their definitions differ slightly between domain adaptation and reference-guided image synthesis due to the varying nature of source and target data, yet both share the core objective of measuring the semantic shifts to enhance diversity and preserve content. CLIP-Across captures the directional relationship between source and target/reference samples to guide adaptation, while CLIP-Within ensures that transformations maintain internal consistency within the adapted domains or transformations. These losses are instrumental in refining the generator's ability to retain identity and style information effectively, aligning generated outputs with the desired target characteristics. We provide graphical illustrations in Fig.~\ref{fig:directional_losses} to clarify their definitions and distinctions across domain adaptation and reference-guided image synthesis tasks.

\section{Evaluation Details}
\label{sec:eval_setup}
\paragraph{Domain Adaptation Experiments}
To quantitatively assess the quality and fidelity of the generated images, we used the widely used Fréchet Inception Distance (FID) score~\cite{FID} and the Quality and Diversity metrics suggested in~\cite{alanov2023hyperdomainnet}. The FID score provides a measure of the statistical distance between the distributions of real and generated images. The Quality metric evaluates how closely the adapted images align with the text description of the target domain. This is computed as the mean cosine similarity between the CLIP embeddings of the images and the CLIP embedding of the text description. The Diversity metric, on the other hand, measures the variability among the adapted images. This is quantified as the mean pairwise cosine distance between the CLIP embeddings of all the adapted images. In our evaluation, we generate a set of 1K images for each target domain using the NADA-expanded Domain Expansion model, treating these images as real. For the FID evaluation, we compare the distribution of these images with that of images generated by the evaluated methods. Specifically, we randomly sample 100 images from each target domain to represent the generated image distribution.

\paragraph{Reference-Guided Image Synthesis Experiments}
%We use FID~\cite{FID}, ID similarity with the source \& target images, CLIP~\cite{radford2021learning} and BLIP~\cite{li2022blip} semantic similarity scores for the quantitative comparisons. 
We use FID~\cite{FID} to measure the quality and the fidelity of the synthesized images as a lower FID score indicates that the synthesized images are closer to the FFHQ domain, which is the original domain the StyleGAN2 is trained on. We use the ID similarity~\cite{Deng_2022} to measure identity preservation. Ideally, we want the ID similarity with the source image to be high and ID similarity with the target image to be low as we want to preserve the identity of the source image while only transferring the attributes of the target image to the source. Finally, we use the CLIP embedding space to measure the semantic similarity of the target and output images to evaluate how well the semantics of the target image are transferred. 

\paragraph{Text-Guided Image Manipulation Experiments}
To evaluate the approaches quantitatively, we employ multiple quantitative metrics, namely Fréchet Inception Distance (FID) \cite{FID}, Attribute Manipulation Accuracy (AMA), and CLIP Manipulative Precision (CMP) following the methodology introduced by CLIPInverter~\cite{CLIPInverter}. FID serves as a measure of the quality and fidelity of the synthesized images. 
 
Attribute Manipulation Accuracy (AMA)~\cite{CLIPInverter} measures how well a single manipulation is applied. To calculate the AMA score of a model, for each attribute (such as \textit{blonde hair}), we first select 50 images that the attribute is not present in. Then, we edit these images with a corresponding caption, such as \textit{The person has blonde hair}. Finally, we use pre-trained attribute classifiers to measure the manipulation accuracy on the output images. We average the accuracy accross the attributes to obtain the final AMA score. We trained attribute classifiers for each of the 40 attributes that are present in the CelebA~\cite{liu2015faceattributes} dataset. We used the 15 attributes that achieve 90\% or higher validation accuracies to calculate the AMA scores. Here is a full list of attributes we used for the CelebA dataset:
\begin{multicols}{2}
\begin{itemize}
    \item \textit{blonde hair}
    \item \textit{bushy eyebrows}
    \item \textit{chubby}
    \item \textit{double chin}
    \item \textit{eyeglasses}
    \item \textit{goatee}
    \item \textit{gray hair}
    \item \textit{heavy makeup}
    \item \textit{male}
    \item \textit{mouth slightly open}
    \item \textit{mustache}
    \item \textit{rosy cheeks}
    \item \textit{smiling}
    \item \textit{wearing lipstick}
    \item \textit{wearing necktie}
\end{itemize}
\end{multicols}
\label{sec:metrics}

In order to quantify the alignment between the output images and the target captions, while preserving the contents of the input image, we employ CMP, which is defined as $\text{CMP} = (1 - \text{diff}) \cdot \text{sim}$, with $\text{diff}$ denoting the L1 pixel difference between the input and output images, and $\text{sim}$ denotes the CLIP semantic similarity between the output image and the target description.

\section{User Study}
\label{sec:user_study}
To further assess our approach and compare it with other competing approached across all three tasks, we conduct a user study using Qualtrics. In the domain adaptation task, we compare our model with DynaGAN~\cite{Kim2022DynaGAN} and HyperDomainNet~\cite{alanov2023hyperdomainnet}, both of which also facilitate adaptation across multiple domains with a single model architecture, akin to ours. For reference-guided image synthesis, our comparisons include TargetCLIP-E~\cite{chefer2021targetclip}, TargetCLIP-O~\cite{chefer2021targetclip}, and BlendGAN~\cite{liu2021blendgan}. For text-guided image manipulation, we evaluate our framework against DeltaEdit~\cite{Lyu_2023_CVPR}, which similarly does not utilize textual textual data during its training phase.

\begin{figure}[!t]
  \centering
  \includegraphics[width=1.0\linewidth]{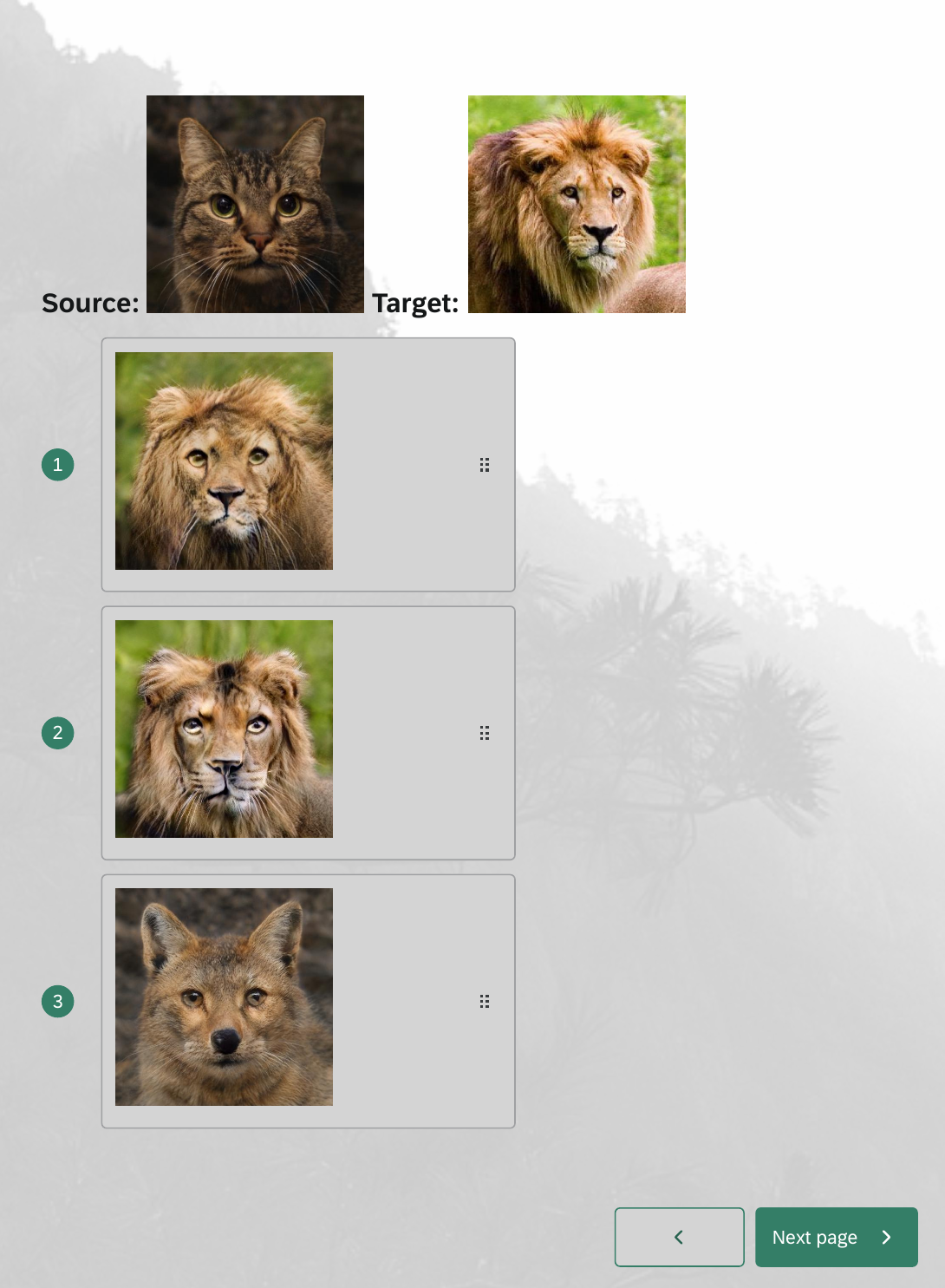}
  \caption{\textbf{A sample question from the user study.} The participants rank the options from best to worst.} 
  \label{fig:user_study_screenshot}
\end{figure}

The human evaluation comprises three sections, each dedicated to one of our tasks, with 25 questions per section. Within each part, users are shown a random source image alongside a target image (or target text in the case of text-guided editing). In the domain adaptation and reference-guided editing sections, participants rank the results based on each model's performance by arranging the images in order of preference, with the top position indicating the best result. For text-guided editing section, participants choose the superior output between our model and DeltaEdit. To mitigate any bias in the evaluation process, the order in which results are displayed is randomized. An example question from the user study is illustrated in Fig.~\ref{fig:user_study_screenshot}.

In Table~\ref{tab:user_study}, we present the average human rankings of the methods for all three tasks. As shown, for text-guided image manipulation, nearly all participants consistently preferred the results of HyperGAN-CLIP over those of DeltaEdit. In the domain adaptation task, HyperGAN-CLIP and DynaGAN received similar rankings, indicating comparable performance. For reference-guided image synthesis, the task appears more subjective, as all average rankings are above 2, with HyperGAN-CLIP showing competitive results against the TargetCLIP models.

\begin{table}[!t]
    \caption{\textbf{User study results.}}

    \begin{tabular}{lc}
        \toprule
             \multicolumn{2}{c}{\textbf{Multiple Domain Adaptation}} \\
             \midrule
            
             Method & Ranking\\
	       \midrule
   		 HyperDomainNet & 2.77\\
		 DynaGAN & 1.58 \\ 
		 Ours & 1.65 \\          \bottomrule\\
         
             \toprule
             \multicolumn{2}{c}{\textbf{Reference-Guided Image Synthesis}} \\
             \midrule
             
             Method & Ranking\\
	       \midrule
          BlendGAN & 3.24\\
          TargetClip-O & 2.20\\
          TargetClip-E & 2.13\\
		 Ours & 2.43 \\          \bottomrule\\
             \toprule
             \multicolumn{2}{c}{\textbf{Text-Guided Image Manipulation}} \\         
             \midrule
             
             Method & Ranking\\
	       \midrule
             DeltaEdit & 1.93\\	 
             Ours & 1.07 \\

         \bottomrule
\end{tabular}
\label{tab:user_study}
\end{table}

\section{Controllable Manipulation} 
\label{sec:controllable_generation}
\begin{figure}[!b]
  \centering
  \includegraphics[width=\linewidth]{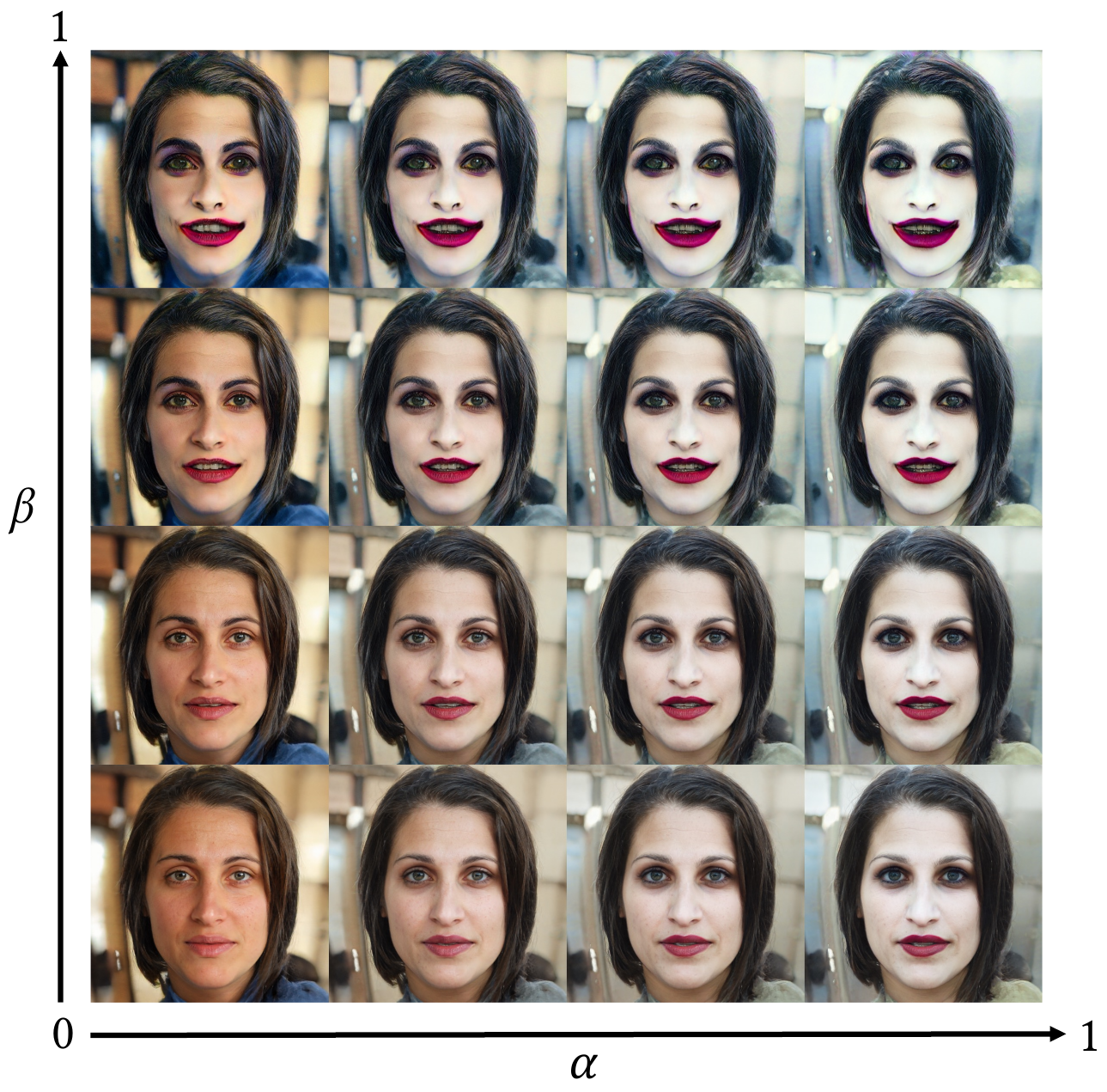}
  \caption{\textbf{Controllable Manipulation.} In our approach, we can vary the amount of residual features injected as well as the amount of target style latent, which gives users the ability to control degree of adaptation with respect to style consistency vs data fidelity.} 
  \label{fig:controllable}
\end{figure}

We observe that scaling the CLIP embeddings translates roughly to scaling of the modulation weights, and consequently, feature maps. By adjusting the scaling ratio of the residual target domain features injected into the source domain features by a factor $\beta$, we can control the degree of adaptation during inference. Furthermore, in line with previous GAN domain adaptation studies, we can enhance the style quality by employing style mixing over the latent codes. This involves interpolating the original latent code $w$ with the target domain’s latent code $w_t$ as $\hat{w} = \alpha * w + (1-\alpha) * w_t$, where $\alpha$ is a scalar between 0 and 1, controlling the level of style mixing. The resulting interpolated latent code $\hat{w}$ can be then used as input to the generator for image synthesis. We demonstrate the impact of these control parameters on the generated images in Fig.~\ref{fig:controllable}.

\begin{figure}[!t]
  \centering
  \includegraphics[width=1.0\linewidth]{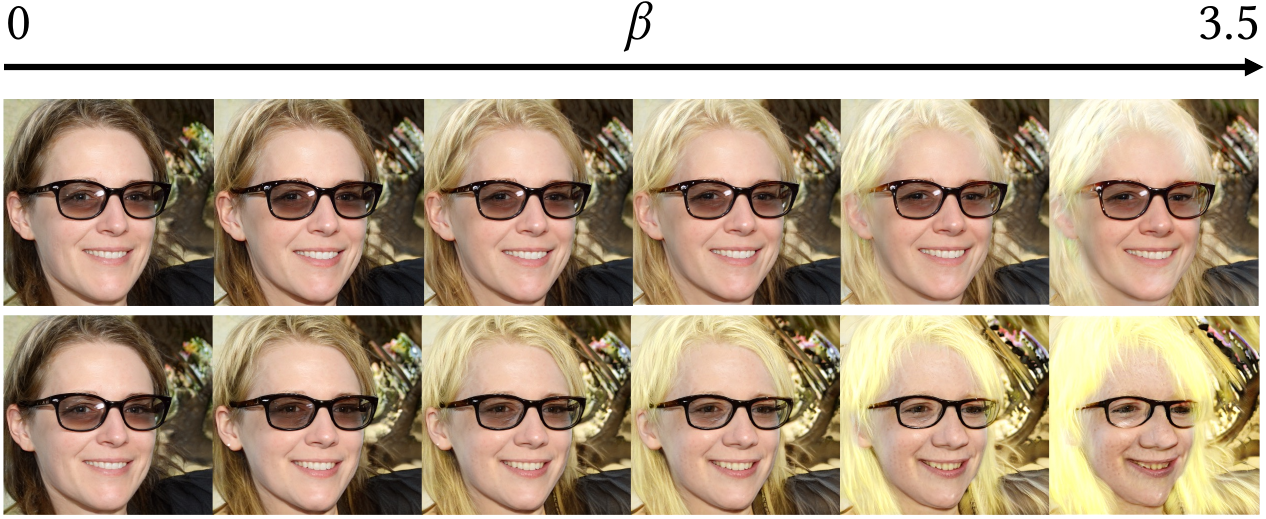}
  \caption{\textbf{Scaling residual features.} The top row shows the results obtained with our approach, whereas the bottom on corresponds to the results by the baseline model with the discriminator. Several artifacts instantly start to appear in the baseline results when scaling beyond the training value of 1 (third column corresponds to $\beta=1$). On the other hand, our approach works relatively seamlessly and preserves the identity better.} 
  \label{fig:controllable_text}
\end{figure}

Moreover, we observe that even if we additionally scale the amount of features injected into the original image features by some factor $\beta$, our approach gives very plausible results. It does not change the features that not related to the target. We compare it with our baseline model with conditional discriminator trained with $\Delta$-CLIP embeddings, where we use the same $\beta$ to scale the modulation parameters to control the degree of adaptation (as in DynaGAN). Since, it does not use original domain features and weight modulation is responsible for preserving both original image characteristics and transferring target style content from CLIP embeddings, it fails to scale well with amount of the feature scaling parameter, as demonstrated in Fig. \ref{fig:controllable_text}. This highlights the importance of using domain specific features as residuals to the original features instead of directly generating the overall combined features.

\section{Zero-Shot Domain Adaptation}
\begin{figure}[!h]
  \centering
\includegraphics[width=0.99\linewidth]{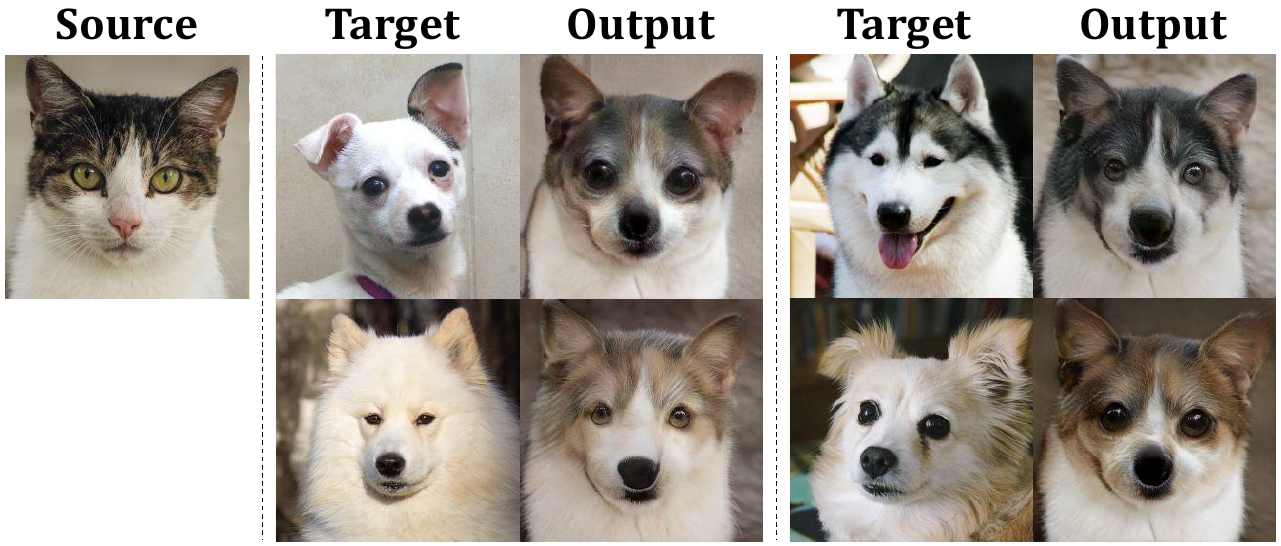}
\caption{\textbf{Zero-Shot Domain Adaptation.} Our model can perform domain adaption quite reasonably well on target domains not seen during training. Here we provide results on target dog breed images from the AFHQ dog dataset not used in the training.}
  \label{fig:zero-shot-domain}
\end{figure}

The use of CLIP conditioning in the design of our proposed hypernetwork module has critical advantages over the prior methods. In this way, during training the model can not only exploit the common characteristics shared among target domains, but it also allows for zero-shot domain adaptation, especially well when the novel target domain not seen during training is semantically close to the target domains in training data. In Fig.\ref{fig:zero-shot-domain}, we provide example results on target dog breed images from the AFHQ dog dataset not used in the training.

\begin{figure*}[!t]
  \centering
  \includegraphics[width=0.99\linewidth]{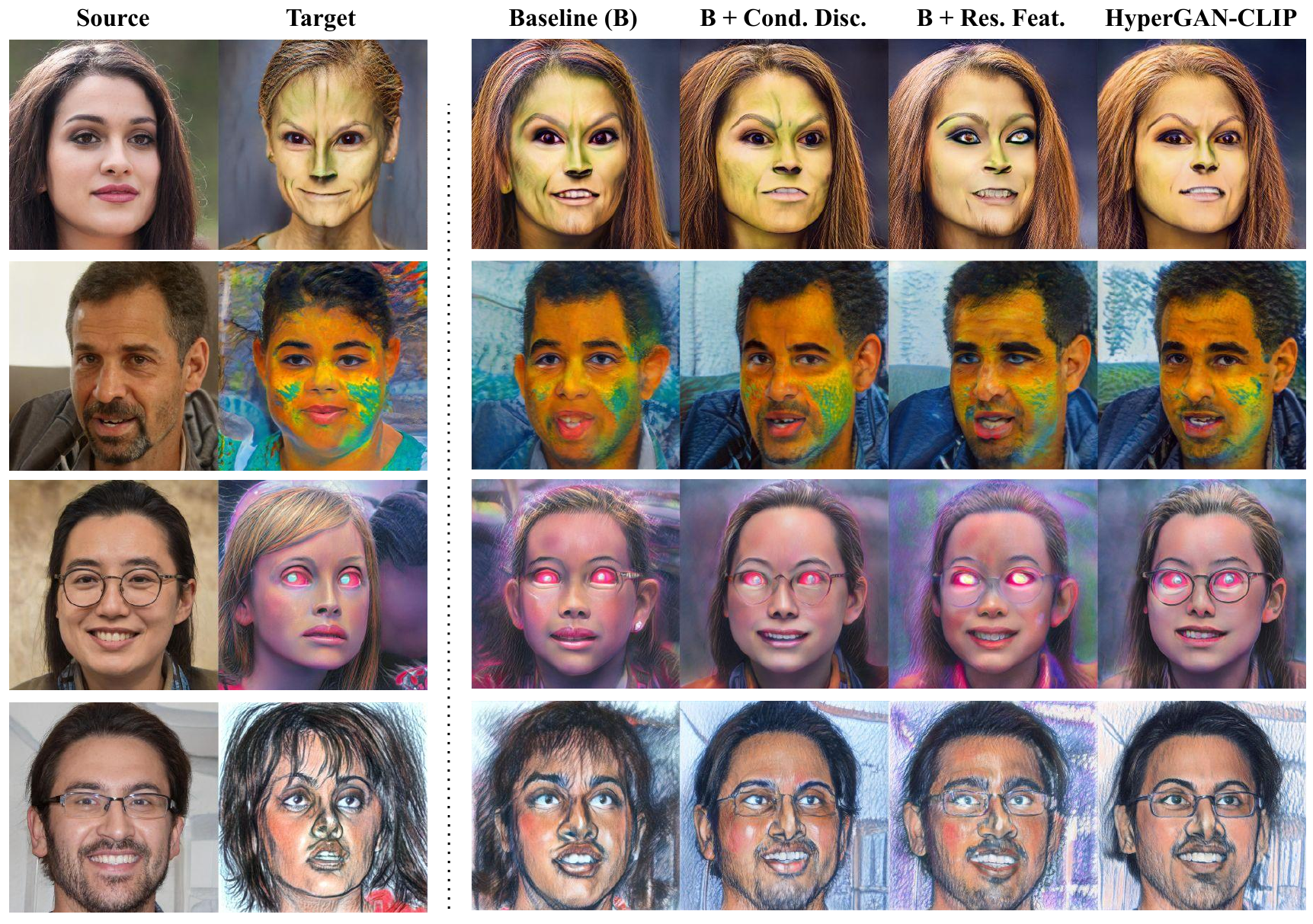}
  \caption{\textbf{Qualitative results for the ablation study.} Baseline network does not preserve the facial identity of the source image, giving an outcome closely resembling to the target image. When CLIP-conditioned discriminator is incorporated to the baseline, the image quality is improved. Using residual features scheme preserves the facial identity better. Our full model gives the best results in terms of both identity and image quality.}
  \label{fig:ablation}
\end{figure*}

\section{Ablation study} 
\label{sec:ablation}
We conduct an ablation study to assess the impact of each component in our model on domain adaptation performance. The qualitative and quantitative results are presented in Fig. \ref{fig:ablation} and Table \ref{tab:ablations_quantitative}, respectively. The baseline network uses only the features given by the target-domain modulated generator, and ignores the source domain features. This approach results in the loss of the source identity and is prone to overfitting to the provided target image. Adding a conditional discriminator loss helps to mitigate the problems to some extent and enhances image quality. Considering residual features scheme that employs target features alongside with the source features preserves the facial identity better than the baseline, but falls short in terms of image quality. Finally, our full model, HyperGAN-CLIP, which utilizes residual feature scheme together with a conditional discriminator effectively preserves identity while capturing target style and maintaining high image quality. 

\begin{table}[!t]
    \caption{\textbf{Quantitative analysis of the ablation study.} The baseline model that employs target-domain modulated features gives the worst score. However, incorporating CLIP-conditioned discriminator and leveraging residual features scheme introduce notable improvements. Our full HyperCLIP-GAN model, utilizing all these components achieves the best score.}
    \centering
    \begin{tabular}{l c}
        \toprule
             Component & \multicolumn{1}{c}{FID~$\downarrow$}\\
        \midrule
        Baseline (B) & 43.43 \\
        Baseline + Cond. Disc. (B + CD) &  33.76\\
        Baseline + Res. Features (B + RF) & 33.43 \\
        HyperGAN-CLIP (B + CD + RF) & \textbf{30.55} \\
        \bottomrule
    \end{tabular}
    \label{tab:ablations_quantitative}
\end{table}

\section{Impact of $\Delta$-CLIP Embeddings}
\label{sec:delta_space}
\begin{figure*}[!t]
  \centering
\includegraphics[width=0.99\linewidth]{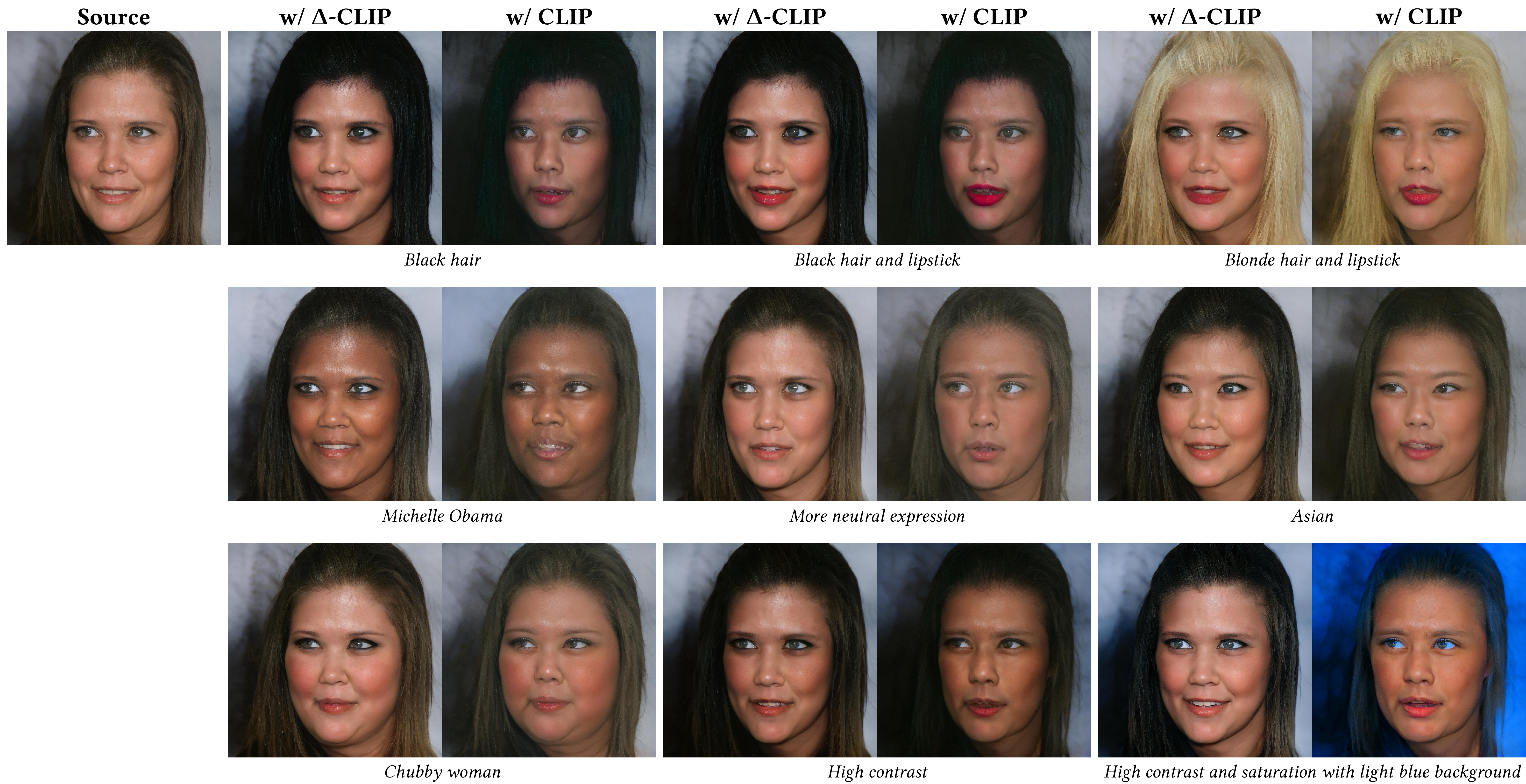}
\caption{\textbf{Impact of $\Delta$-CLIP Embeddings.} Our model equipped with $\Delta$-CLIP embeddings performs semantic edits that are better aligned with the provided textual descriptions as compared to the version of our model that employs original CLIP embeddings.}
  \label{fig:deltaCLP_qualitative}
\end{figure*}

\begin{figure*}[!t]
  \centering
  \includegraphics[width=0.99\linewidth]{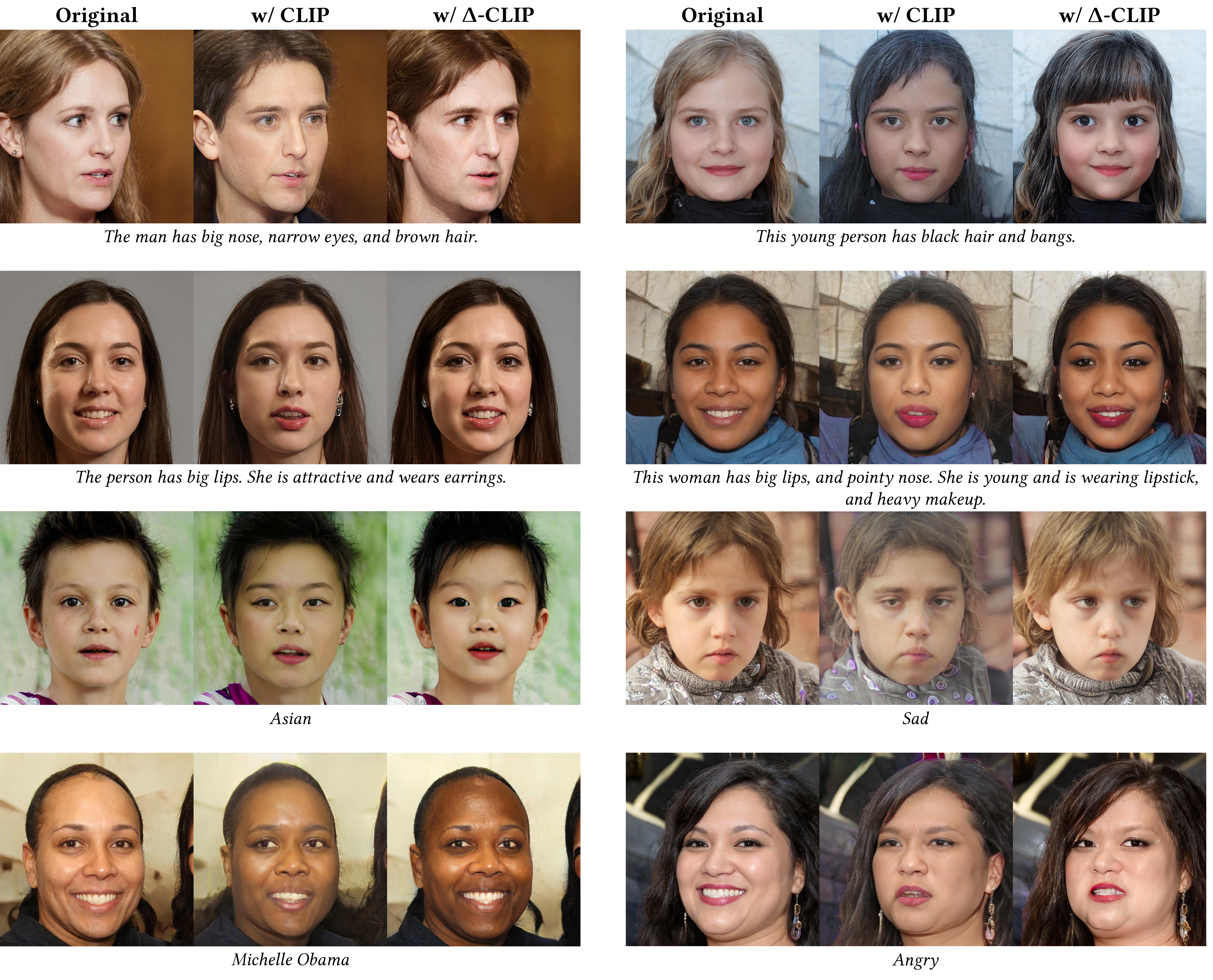}
  \caption{\textbf{Impact of $\Delta$-CLIP Embeddings.} As observed, we obtain much accurate manipulations while preserving the quality and fidelity when $\Delta$-CLIP embeddings are used. }
  \label{fig:supp_delta_space}
\end{figure*}

As stated in the main paper, the $\Delta$-CLIP space provides a semantic embedding space that offers improved alignment between text and image modalities compared to the original CLIP space. This distinction becomes particularly evident when examining the text-guided image manipulation task. In Fig.~\ref{fig:deltaCLP_qualitative} and Fig.~\ref{fig:supp_delta_space}, we show the impact of utilizing these spaces within our hypernetworks module adjusting the weights of the pre-trained StyleGAN generator. The results demonstrate that the $\Delta$-CLIP embeddings enable highly precise text-based editing while preserving image quality and identity fidelity.

\section{Additional Performance Comparisons}
\label{sec:additional_results}
\begin{figure*}[!t]
  \centering
  \includegraphics[width=\linewidth]{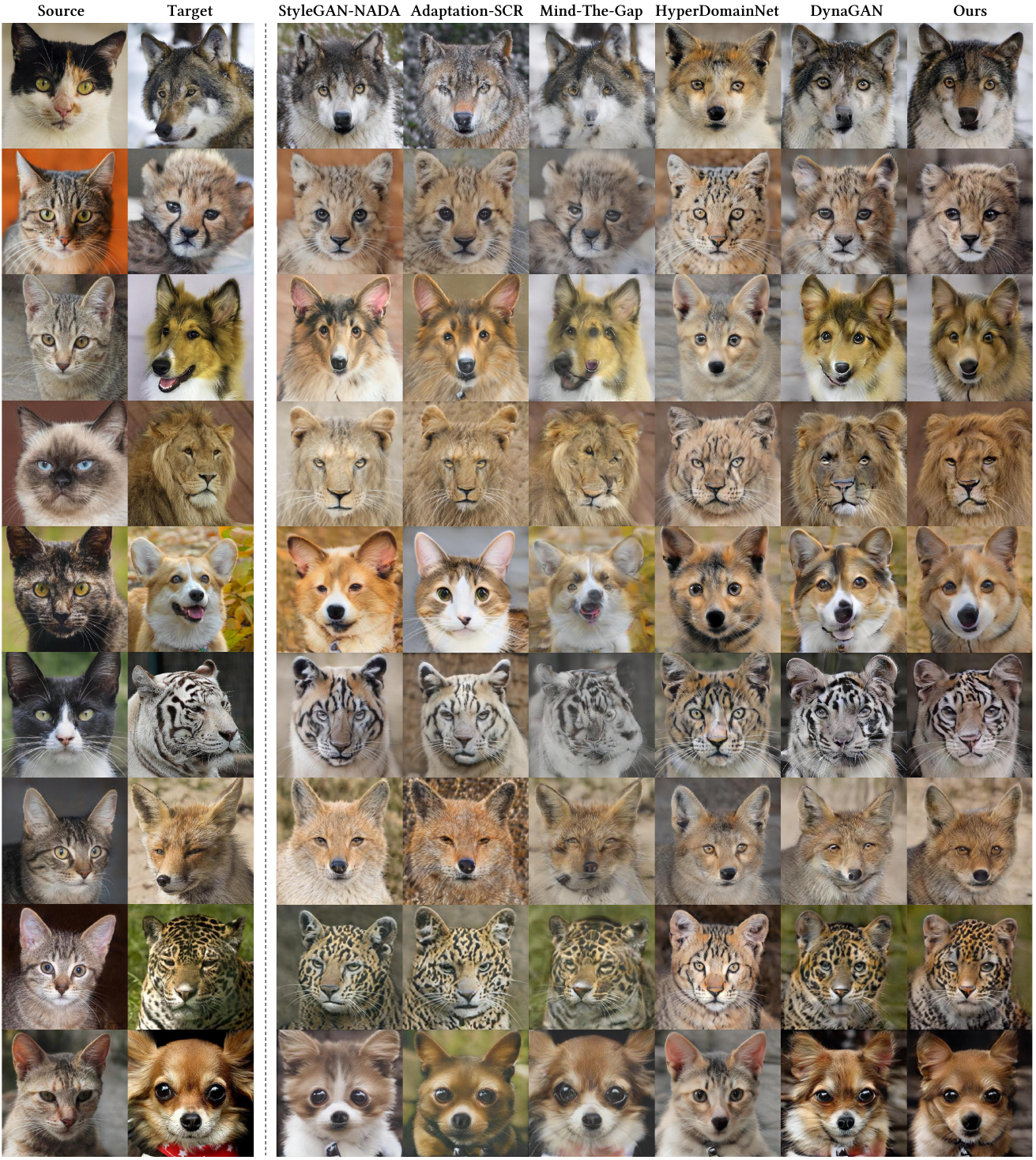}
  \caption{\textbf{Additional qualitative comparison against the state-of-the-art few-shot domain adaptation methods on AFHQ dataset.} Our proposed HyperGAN-CLIP model outperforms competing methods in accurately capturing the visual characteristics of the target domains. The synthesized images exhibit a higher degree of fidelity and realism, demonstrating the effectiveness of our approach.
  }
  \label{fig:supp_domain_adaptation1}
\end{figure*}

\begin{figure*}[!t]
  \centering
  \includegraphics[width=\linewidth]{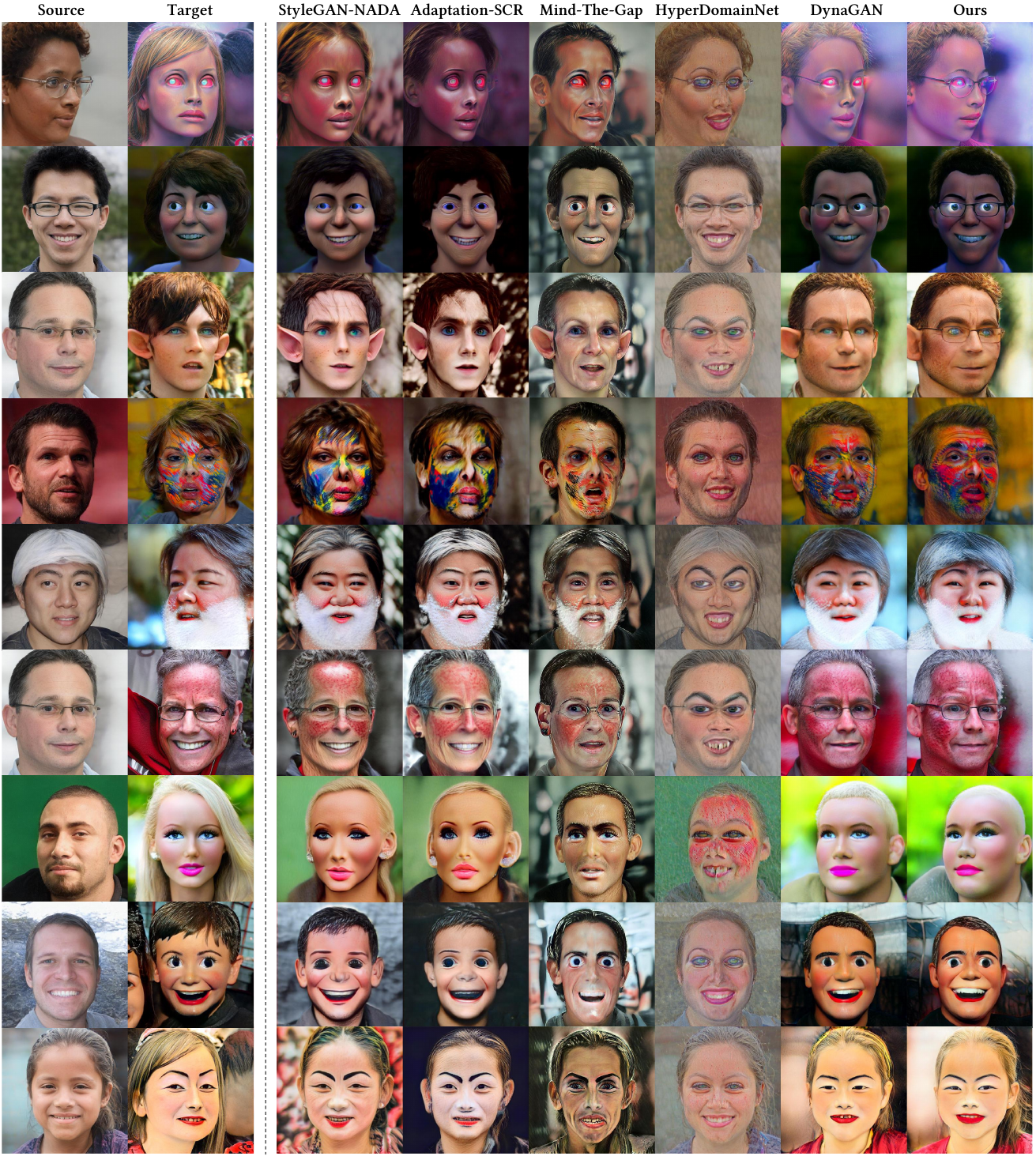}
  \caption{\textbf{Additional qualitative comparison against the state-of-the-art few-shot domain adaptation methods on AFHQ dataset.} Our proposed HyperGAN-CLIP model outperforms competing methods in accurately capturing the visual characteristics of the target domains. The synthesized images exhibit a higher degree of fidelity and realism, demonstrating the effectiveness of our approach.
  }
  \label{fig:supp_domain_adaptation2}
\end{figure*}

\begin{figure*}[!t]
  \centering
  \includegraphics[width=0.95\linewidth]{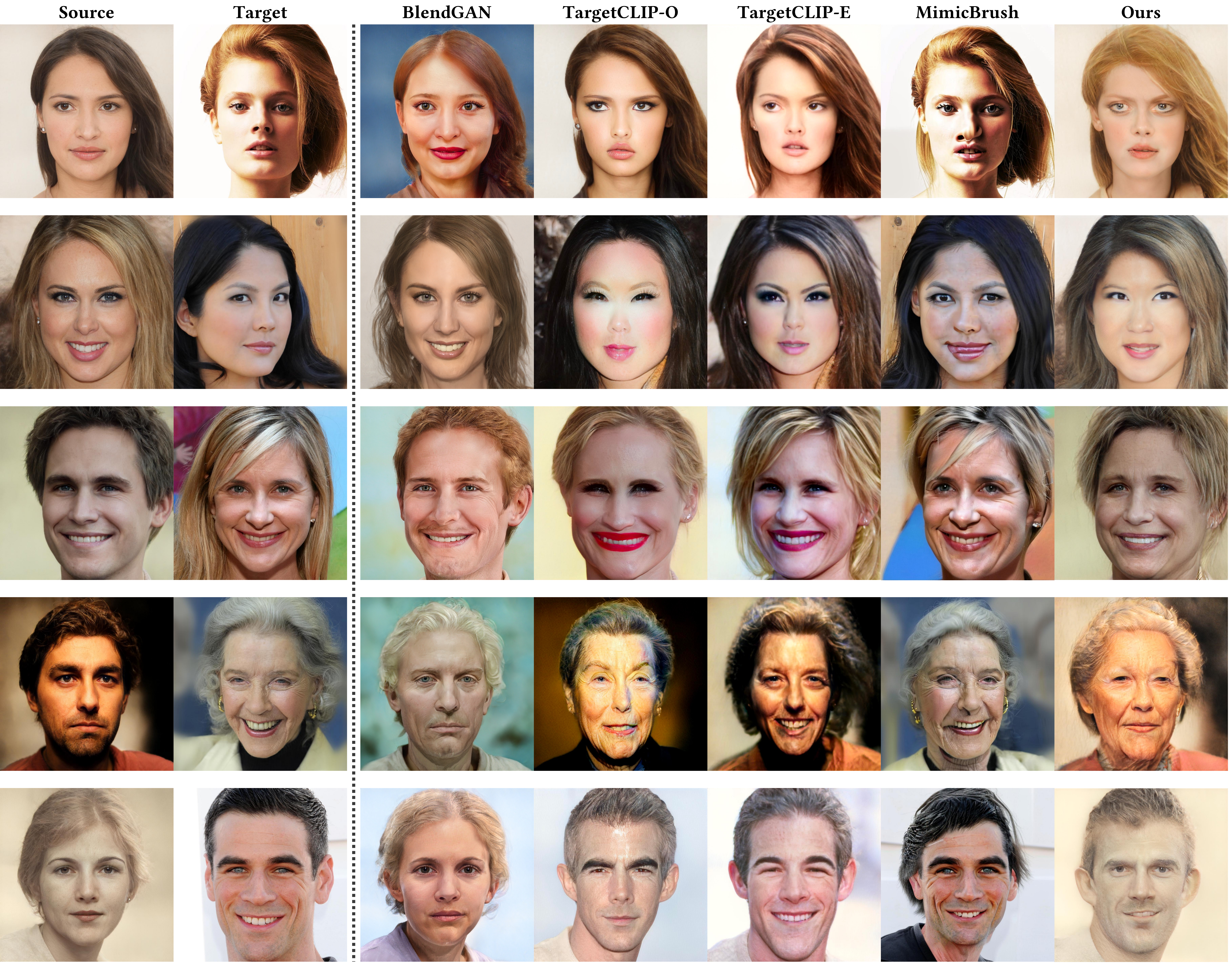}
  \caption{\textbf{Additional qualitative comparison with state-of-the-art reference-guided image synthesis approaches.} Our approach effectively transfers the style of the target image to the source image while effectively preserving identity compared to competing methods.}
  \label{fig:supp_image_based}
\end{figure*}

\begin{figure*}[!t]
  \centering
  \includegraphics[width=0.99\linewidth]{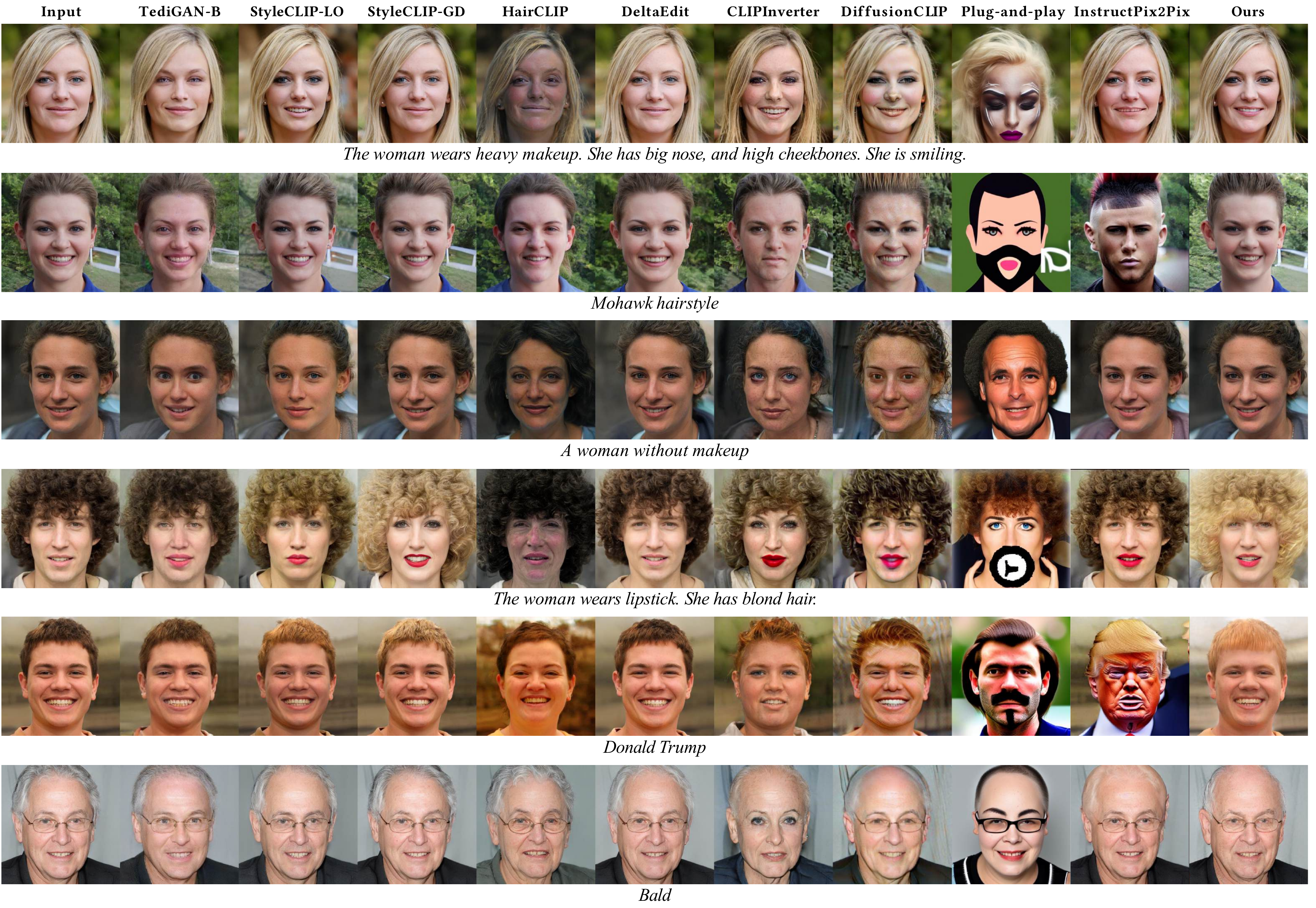}
  \caption{\textbf{Additional qualitative comparisons with state-of-the-art text-guided image manipulation methods.} Our model shows remarkable versality in manipulating images across a diverse range of textual descriptions. The results vividly illustrate our model's ability to accurately apply changes based on target descriptions encompassing both single and multiple attributes. Compared to the competing approaches, our model preserves the identity of the input much better while successfully executing the desired manipulations.
  }
  \label{fig:supp_text_based}
\end{figure*}

In Fig.~\ref{fig:supp_domain_adaptation1} and Fig.~\ref{fig:supp_domain_adaptation2}, we present additional comparisons in multple domain adaptation against Mind-the-GAP~\cite{zhu2021mind}, StyleGAN-NADA~\cite{gal2021stylegannada}, HyperDomainNet~\cite{alanov2023hyperdomainnet}, DynaGAN~\cite{Kim2022DynaGAN}, and Adaptation-SCR~\cite{Liu2023scr} on AFHQ and FFHQ datasets, respectively. In Fig.~\ref{fig:supp_image_based}, we give additional comparisons of our approach against the BlendGAN~\cite{liu2021blendgan}, TargetCLIP-O~\cite{chefer2021targetclip} and TargetCLIP-E~\cite{chefer2021targetclip} models in reference-guided editing. Finally, in Fig.~\ref{fig:supp_text_based}, we provide additional qualitative comparisons in text-driven manipulation against TediGAN-B~\cite{xia2021tedigan}, StyleCLIP-LO~\cite{Patashnik_2021_ICCV}, StyleCLIP-GD~\cite{Patashnik_2021_ICCV}, HairCLIP~\cite{wei2022hairclip}, DeltaEdit~\cite{Lyu_2023_CVPR}, CLIPInverter~\cite{CLIPInverter}, DiffusionCLIP~\cite{Kim_2022_CVPR} and plug-and-play~\cite{pnpDiffusion2022}.

\begin{figure*}[!t]
  \centering
  \includegraphics[width=0.99\linewidth]{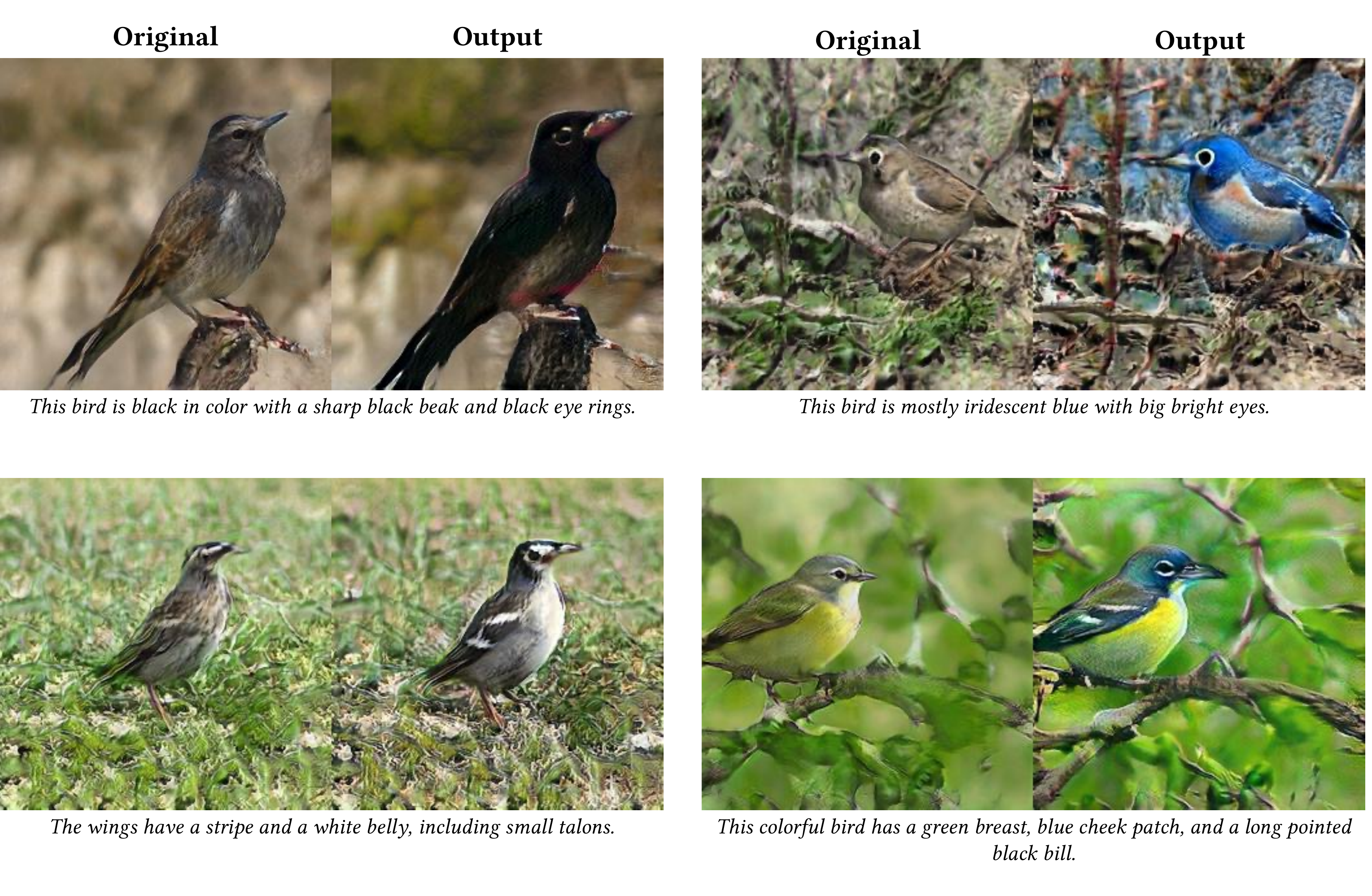}
  \caption{\textbf{Text-guided editing results for the birds dataset.} Our approach generalizes to other domains, such as the bird images. We demonstrate zero-shot text-guided image editing results.}
  \label{fig:supp_bird_text}
\end{figure*}

\begin{figure*}[!t]
  \centering
  \includegraphics[width=0.99\linewidth]{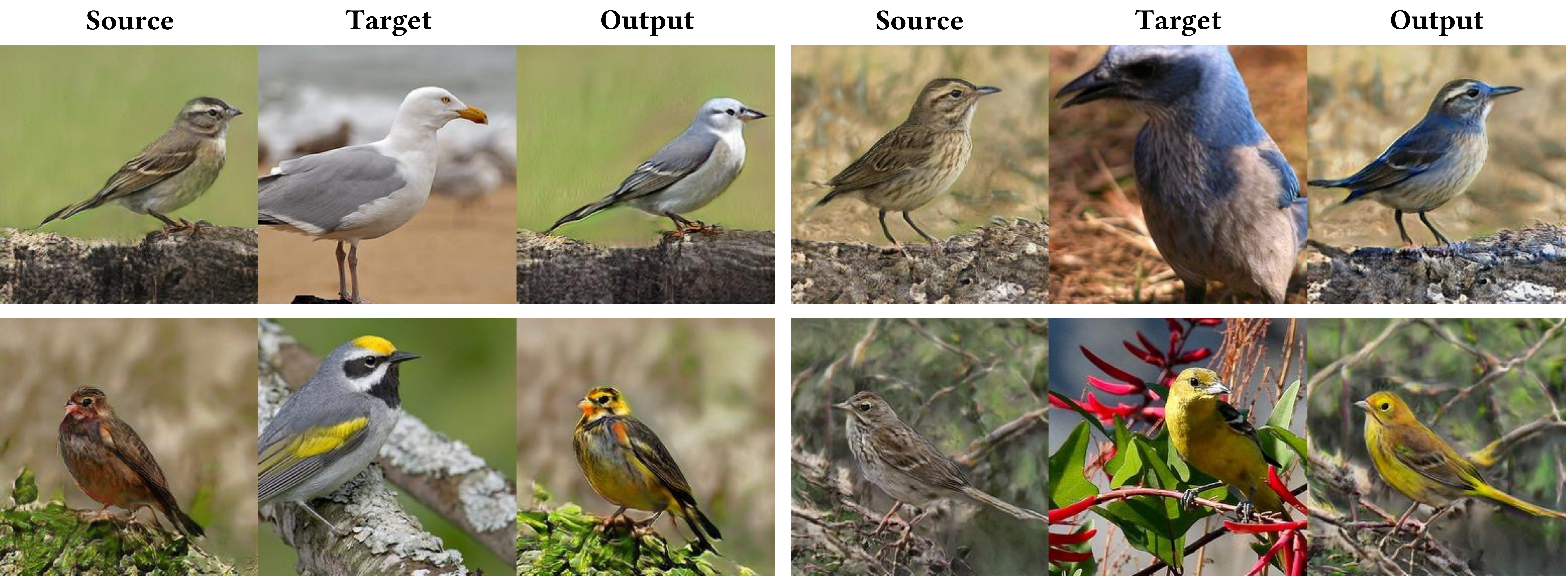}
  \caption{\textbf{Reference-guided editing results for the birds dataset.} Our reference-guided synthesis generalizes to the birds domain, illustrated by the various targets we provide.}
  \label{fig:supp_bird_image}
\end{figure*}

We trained our HyperCLIP-GAN on the CUB-Birds dataset~\cite{WahCUB_200_2011} as well to demonstrate the generalization capabilities of our approach. When training these models, we use the same losses as described in the main paper except for the identity preservation loss, where we alternatively employ a ResNet50~\cite{he2015deep} network trained with MOCOv2~\cite{chen2020improved}. In Fig.~\ref{fig:supp_bird_text}, we provide various text-guided editing results, and in Fig.~\ref{fig:supp_bird_image}, we provide several reference-guided synthesis results for the CUB-Birds dataset.

\begin{figure*}[!t]
  \centering
  \includegraphics[width=0.87\linewidth]{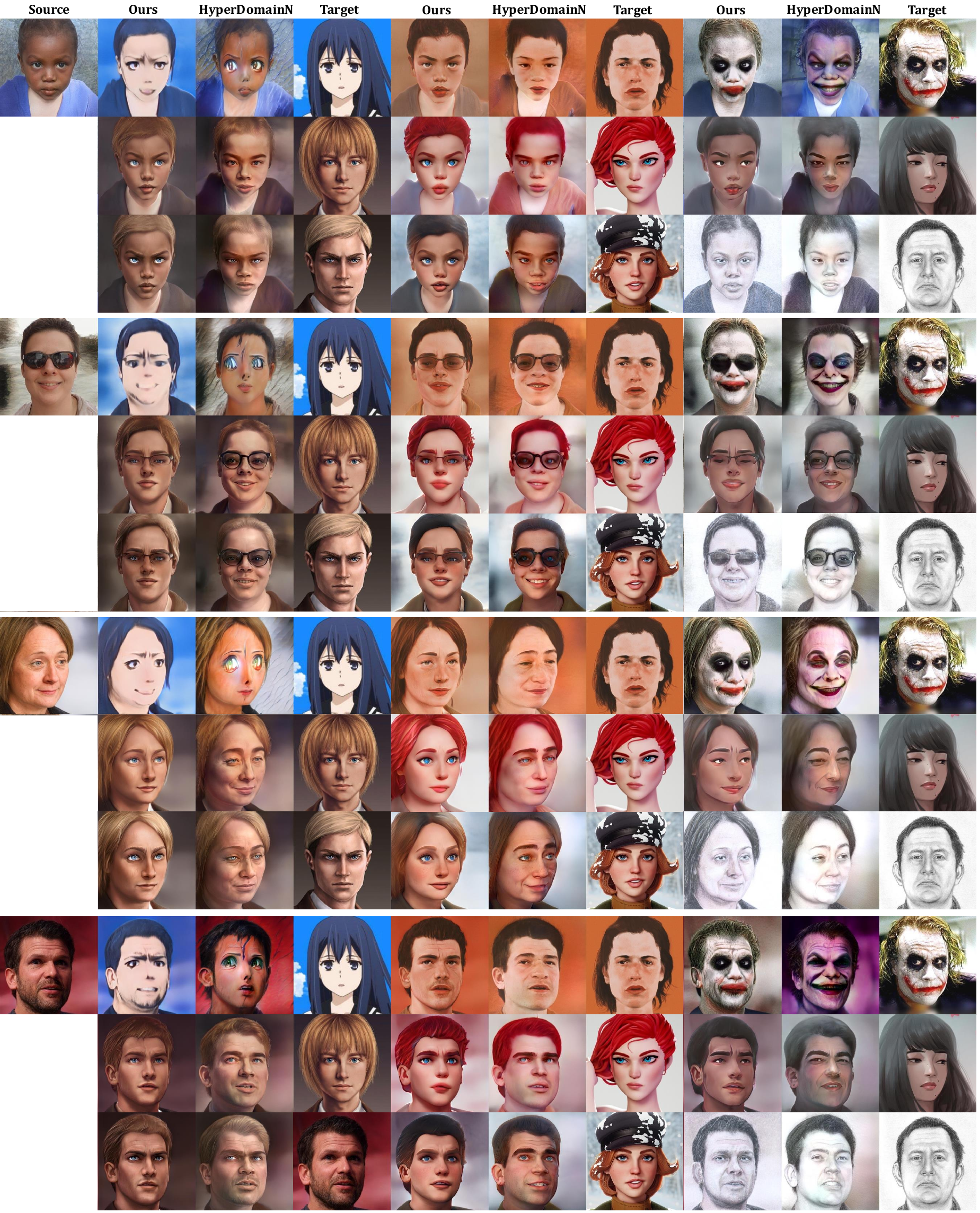} 
  \caption{\textbf{Additional qualitative comparison with HyperDomainNet.} The comparisons on a smaller set of domains shows that our proposed HyperGAN-CLIP model performs comparably or better than HyperDomainNet.}
  \label{fig:hdn_comparison}
\end{figure*}

We also perform an additional analysis covering our approach and HyperDomainNet, two hypernetworks-based multi-domain adaptation approaches. In particular, we train our framework on a much smaller, less diverse set of domains involving one image per domain and consisting of 20 different domains from~\cite{alanov2023hyperdomainnet}. In Fig.~\ref{fig:hdn_comparison}, we provide sample side-by-side qualitative comparisons using the pre-trained model provided by the authors. Overall, the results show that our method performs either better or on par with HyperDomainNet on this more limited set of domains.

\section{Limitations}
\label{sec:limitations}
HyperGAN-CLIP performs optimally when the target domain shares a resemblance with the source domain in terms of content. However, when there is a significant domain gap, it struggles to adapt the pre-trained generator to the target domain. Additionally, for reference-guided image synthesis and text-guided image manipulation, HyperGAN-CLIP can produce visually plausible results only for concepts  encountered during training or those that are semantically similar. It fails to generalize to entirely different, unseen concepts.

\section{Ethical Statement}
\label{sec:ethical_statement}
The transformative capabilities of Generative Adversarial Networks (GANs) in image editing, particularly in the realm of human face manipulation present not only technological advancements but also ethical implications that merit careful consideration. Prominent concerns involve the potential for the creation of deceptive or harmful content, exemplified by the emergence of deepfakes, which can be exploited for malicious purposes such as disseminating misinformation or perpetuating identity theft. Moreover, biases encoded in training data may perpetuate societal prejudices. This study emphasizes responsible research practices, advocating for transparent disclosure of limitations and risks. Open dialogue within the research community is crucial for addressing these ethical implications. Implementing safeguards, including content detection methods and adherence to ethical guidelines, is essential for the responsible development and deployment of face editing technologies, ensuring a positive societal impact.

\bibliographystyle{ACM-Reference-Format}
\bibliography{sample-bibliography}